%% file: main.tex
\documentclass{article} 
\usepackage{iclr2026_conference,times}
\input{math_commands.tex}

\usepackage{hyperref}
\usepackage{url}
\usepackage{booktabs}       
\usepackage{amsfonts}       
\usepackage{amsmath}
\usepackage{multirow}    
\usepackage{pifont}      
\usepackage{tablefootnote} 
\usepackage{enumitem}
\usepackage{graphicx}
\usepackage{subfigure}
\usepackage{algorithm}
\usepackage{algpseudocode}
\usepackage{array} 
\usepackage{siunitx} 
\usepackage{wrapfig}

\usepackage{amssymb}
\usepackage[table]{xcolor}
\usepackage[most]{tcolorbox}

\title{DeepSearch: Overcome the Bottleneck of Reinforcement Learning with Verifiable Rewards via Tree-based Search}
\author{\textbf{Fang Wu}$^{\heartsuit}$\thanks{Equal contributions.} \quad
\textbf{Weihao Xuan}$^{\nabla,\triangle}$\footnotemark[1] \quad
\textbf{Heli Qi}$^{\triangle}$\footnotemark[1] \quad 
\textbf{Aaron Tu}$^{\spadesuit}$ \quad  
\textbf{Ximing Lu}$^{\diamondsuit}$ \quad  \\
\textbf{Li Erran Li}$^{\clubsuit}$ \quad  
\textbf{Yejin Choi}$^{\heartsuit}$\thanks{Corresponding author. Email: \texttt{yejinc@stanford.edu}} \\
$^{\heartsuit}$Stanford University\,
 $^{\nabla}$University of Tokyo\,
 $^{\triangle}$RIKEN AIP\,
 $^{\spadesuit}$UC Berkeley \, \\
 $^{\diamondsuit}$University of Washington\, 
 $^{\clubsuit}$Amazon AWS\,
}

\newcommand{\rlvrbold}[1]{\textcolor{green}{\textbf{#1}}} 
\newcommand{\basebold}[1]{\textcolor{red}{\textbf{#1}}}   

\newtcolorbox{casebox}[1][]{
  enhanced,
  breakable,
  colback=gray!5,
  colframe=black!20,
  coltitle=black,
  boxrule=0.5pt,
  arc=2mm,
  left=4mm,
  right=4mm,
  top=2mm,
  bottom=2mm,
  before skip=8pt,
  after skip=8pt,
  #1
}

\iclrfinalcopy 
\begin{document}
\maketitle

\begin{abstract}
    Although Reinforcement Learning with Verifiable Rewards (RLVR) has become an essential component for developing advanced reasoning skills in language models, contemporary studies have documented training plateaus after thousands of optimization steps, i.e., notable decreases in performance gains despite increased computational investment. This limitation stems from the sparse exploration patterns inherent in current RLVR practices, where models rely on limited rollouts that often miss critical reasoning paths and fail to provide systematic coverage of the solution space. We present DeepSearch, a framework that integrates Monte Carlo Tree Search (MCTS) directly into RLVR training. In contrast to existing methods that rely on tree search only at inference, DeepSearch embeds structured search into the training loop, enabling systematic exploration and fine-grained credit assignment across reasoning steps. Through training-time exploration, DeepSearch addresses the fundamental bottleneck of insufficient exploration, which leads to diminishing performance gains over prolonged training. Our contributions include: (1) a global frontier selection strategy that prioritizes promising nodes across the search tree, (2) selection with entropy-based guidance that identifies confident paths for supervision, and (3) adaptive replay buffer training with solution caching for efficiency. Experiments on mathematical reasoning benchmarks show that DeepSearch achieves an average accuracy of 62.95\% and establishes a new state-of-the-art reasoning model, while using 5.7x fewer GPU hours than extended training approaches. These results highlight the importance of strategic exploration over brute-force scaling and demonstrate the promise of algorithmic innovation for advancing RLVR methodologies. DeepSearch establishes a new direction for scaling reasoning capabilities through systematic search rather than prolonged computation. 
\end{abstract}

\begin{center}
\vspace{-1em}
    \includegraphics[height=1em]{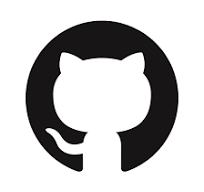}~\url{https://github.com/smiles724/DeepSearch} 
\end{center}
\vspace{-1em}
\begin{center}
    \includegraphics[height=0.85em]{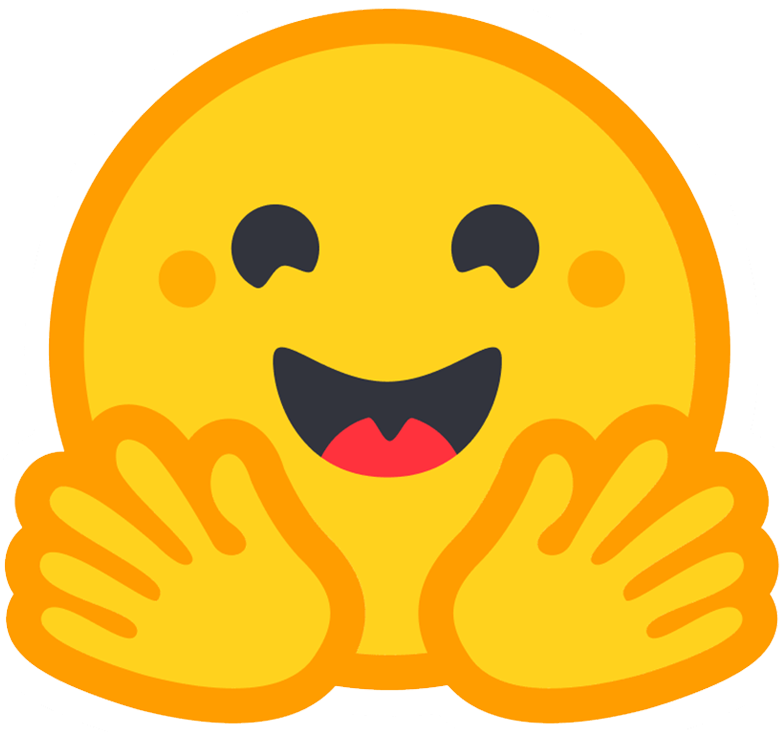}~\url{https://huggingface.co/fangwu97/DeepSearch-1.5B}
\end{center}

\section{Introduction}
Large language models (LLMs) have recently achieved notable progress on complex reasoning tasks~\citep{deepseekai2025deepseekr1incentivizingreasoningcapability,yang2024qwen25mathtechnicalreportmathematical,wu2025large,wang2026reasoning,xia2025agent0}, driven in part by test-time computation scaling strategies~\citep{li2023making,yao2023tree,bi2024forest,zhang2024rest,guan2025rstar} such as tree search with process-level evaluation. While effective, these methods typically treat structured search as an inference-only mechanism, leaving untapped potential to integrate systematic exploration into the training process itself.

This separation between training and inference imposes fundamental limitations on the scalability of reinforcement learning with verifiable rewards (RLVR) for reasoning. Current RLVR approaches remain constrained by sparse exploration patterns during training~\citep{wu2025invisible,liu2025part,tu2025position}, while models are expected to demonstrate sophisticated search behaviors only at inference time. Even recent advances in prolonged RL training~\citep{liu2025prorl} have shown that performance plateaus after thousands of steps, with diminishing returns to allocating more compute to deeper training. This suggests that simply scaling the number of training steps, the primary axis explored in prior work, may not be sufficient to fully realize RLVR's potential.

We address this gap by introducing DeepSearch, a framework that embeds Monte Carlo Tree Search (MCTS)~\citep{metropolis1949monte} directly into RLVR training, representing a fundamental shift from scaling training depth to scaling training breadth. By coupling structured search with verifiable rewards during training, DeepSearch enables models to learn not only from correct solutions but also from the systematic exploration process itself, providing richer supervision than outcome-based or direct rollout methods~\citep{lyu2025exploring,he2025response}.

The core insight driving us is to \emph{focus on training-time exploration} as the driver of improved reasoning.
While traditional RLVR relies on limited rollouts that may miss critical reasoning paths, DeepSearch systematically expands the reasoning frontier during training through principled tree search. This design advances three key objectives: \emph{(i)} expanding reasoning coverage beyond what direct policy rollouts can achieve, \emph{(ii)} providing fine-grained credit assignment to intermediate reasoning steps through tree-structured backpropagation, and \emph{(iii)} maintaining computational efficiency through intelligent node selection and solution caching strategies.

Towards these goals, DeepSearch introduces several key innovations. First, \emph{global frontier selection} strategy prioritizes the most promising nodes across the entire search tree, moving beyond traditional root-to-leaf Upper Confidence Bounds for Trees (UCT) traversals that can be computationally wasteful and myopic. Second, \emph{selection with entropy-based guidance} systematically identifies confident incorrect reasoning paths for supervision. Finally, an adaptive training strategy with replay buffers progressively filters challenging problems and caches verified solutions, thereby avoiding redundant computation across training iterations.

We evaluate DeepSearch on math reasoning benchmarks, where it significantly outperforms strong RLVR baselines~\citep{liu2025prorl,deepscaler2025}. Our results show that DeepSearch achieves 62.95\% average accuracy on challenging mathematical tasks, representing \textbf{a new state-of-the-art for 1.5B reasoning models}. Importantly, these gains are achieved while maintaining computational efficiency through progressive filtering and intelligent reuse of solutions, demonstrating that search-augmented training can be both more effective and more practical than conventional approaches.

The implications extend beyond math reasoning: by bridging the gap between inference-time search capabilities and training-time learning, DeepSearch establishes a new approach for scaling RLVR that emphasizes systematic exploration over prolonged training. This work suggests that the future of reasoning model development lies not just in scaling model parameters or training steps, but in fundamentally rethinking how we structure the learning process to mirror the sophisticated reasoning patterns we expect at inference time. 

\begin{figure}[t]
    \centering
    \includegraphics[width=1.0\linewidth]{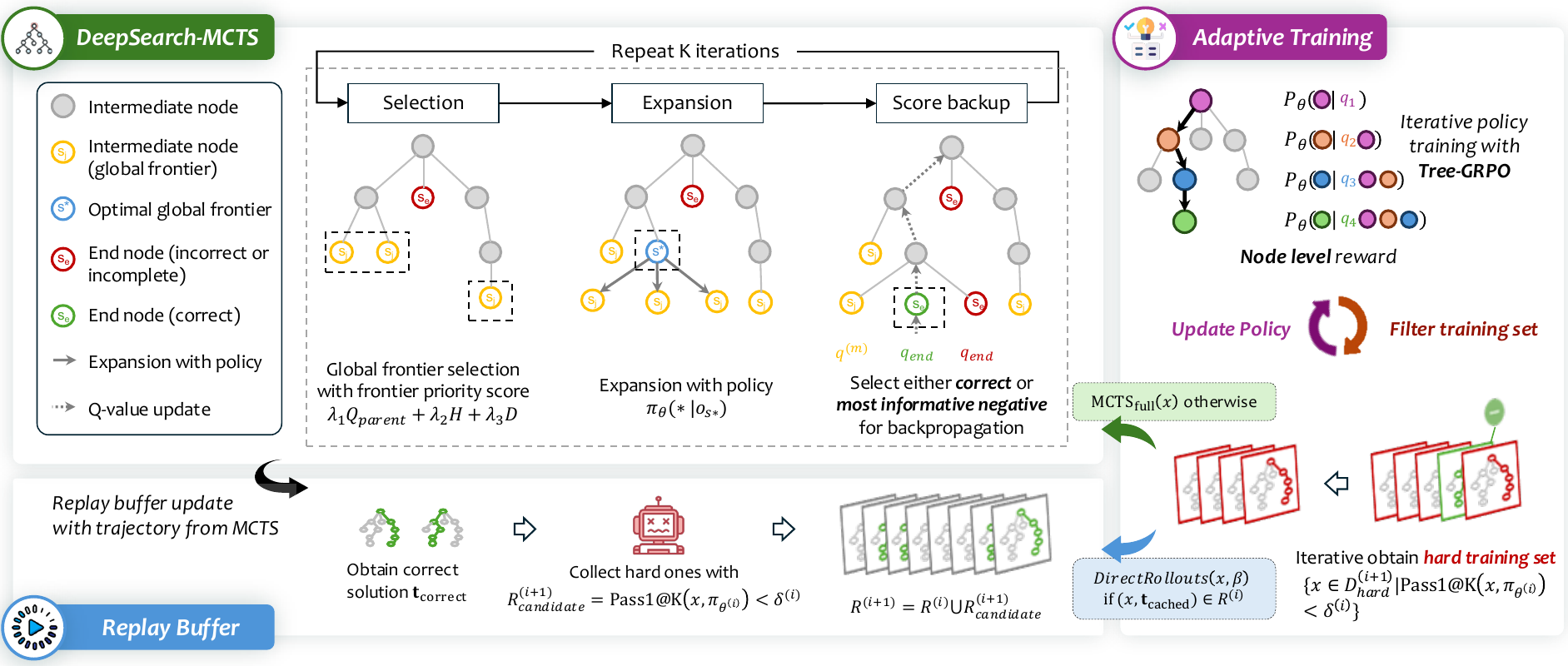}
    \vspace{-2em}
    \caption{DeepSearch framework overview with three key components. }
    \vspace{-1em}
    \label{fig:ds_mcts}
\end{figure}
\vspace{-5mm}

{\section{Related Works}
\label{app:related_work}
\paragraph{Search-based reasoning.}
Structured search has become a standard strategy for scaling test-time compute in LLMs~\citep{snell2024scaling,wu2024inference,zhang2024llama}, with diverse methods including tree-based~\citep{yao2023tree,zhang2024accessing,qi2024mutual} and random sampling approaches~\citep{wang2022self}. Recently, search-based reasoning has evolved into sophisticated frameworks that integrate three core components: policy models, reward models, and search algorithms. Drawing inspiration from game-playing systems like AlphaGo~\citep{silver2016mastering}, works have explored Monte Carlo Tree Search (MCTS) and beam search to guide LLMs through structured reasoning processes~\citep{chen2024alphamath,zhang2024rest,zhang2024llama}, particularly following OpenAI's o1 release~\citep{jaech2024openai}. These frameworks enable exploration of multiple solution paths during inference, trading compute resources for improved accuracy on challenging tasks such as math reasoning. Key design considerations include outcome-supervised versus process-supervised reward models, discriminative versus generative reward architectures, and search strategies ranging from local selection to global exploration~\citep{lightman2023let,wang2023math}. However, most methods restrict search to inference and do not integrate exploration signals into training, leaving the potential for jointly optimizing search and learning largely unexplored.}
\vspace{-1em}

{\paragraph{Reinforcement learning from verifiable rewards.}
RLVR has emerged as a transformative approach for aligning and enhancing LLMs by addressing critical challenges across instruction following~\citep{su2025crossing,gunjal2025rubrics}, ethical alignment~\citep{wang2025rlver}, and reasoning capabilities~\citep{wang2025reinforcement}.
Recent extensions~\citep{guo2025deepseek,yu2025dapo,wan2025srpo} have improved training stability and efficiency by incorporating critic-free optimization, dynamic sampling, and adaptive weighting mechanisms. While these approaches demonstrate the promise of RLVR, they predominantly rely on direct rollouts, which can constrain systematic exploration of the solution space~\citep{wu2025invisible,yue2025does}.}
\vspace{-1em}

{\paragraph{Monte Carlo Tree Search.} MCTS is a powerful search paradigm for complex decision-making problems, extensively explored across diverse fields like games~\citep{silver2016mastering,ye2021mastering}, robotics~\citep{best2019dec,dam2022monte}, theorem proving~\citep{lample2022hypertree}, and matrix multiplication~\citep{fawzi2022discovering}. Early work, such as AlphaGo, integrated MCTS with deep learning~\citep{kemmerling2023beyond}, achieving superhuman performance in board and video games~\citep{ye2021mastering}. Recently, MCTS has been applied to path finding and train timetabling problems~\citep{pitanov2023monte,yang2023integrated}, while~\citet{vagadia2024phyplan} integrated MCTS into physics-informed planning networks for robot control. Despite the demonstrated potential of MCTS for heuristic exploration, it remains unclear how to effectively employ it during RLVR training.}
\vspace{-0.5em}

\section{DeepSearch with MCTS}
Given a problem $x$ and a policy model $\pi_\theta$, we adopt a modified MCTS framework to build a search tree for incremental step-by-step solution exploration. We replace traditional root-to-leaf selection with global frontier-based node selection. The root node represents the question $x$, and child nodes correspond to intermediate steps $s$ generated by $\pi_\theta$. A root-to-leaf path ending at a terminal node $s_\text{end}$ forms a trajectory $\mathbf{t} = x \oplus s_1 \oplus s_2 \oplus \ldots \oplus s_\text{end}$, where each step $s_i$ is assigned a q-value $q(s_i)$. Then we extract solution trajectories $\mathbb{T} = \left\{\mathbf{t}^1, \mathbf{t}^2, \ldots, \mathbf{t}^n\right\}(n \geq 1)$ from the search tree $\mathcal{T}$, where $\mathbf{t}^i$ can be correct, incorrect or incomplete. The depth of any node $s$ is denoted as $d(s)\in \mathbb{Z}^+$. $N(s)$ and $\xi(s)$ denote the number of visits to $s$ and the number of children nodes of $s$, respectively.
Starting from the root node $x$, our MCTS iterations are conducted through three subsequent components. 

\subsection{Expansion with Entropy-based Guidance}

In step $i$, we collect the latest reasoning trajectory $o_i = x \oplus s_1 \oplus s_2 \oplus \ldots \oplus s_{i-1}$ as the current state, i.e., observation. Based on this state, we prompt the policy model $\pi_\theta(s_i|o_i)$ to generate $n$ candidates for the next-step reasoning trail $\{s_{i,j}\}_{j=1}^n$. We repeat this expansion behavior until we reach the terminal nodes $s_\text{end} \in \mathcal{S}_\text{end}$, either by arriving at the final answers or by hitting the maximum depth of the tree $d_\mathcal{T}$, which yields an ordered sequence $s_1\rightarrow \cdots \rightarrow s_\text{end}$.

During each expansion, let $\mathcal{S}_\text{end}^{(k)}$ denote the set of newly generated terminal nodes at iteration $k$. We evaluate the correctness of each terminal node using a verification function $\mathcal{V}: \mathcal{S}_\text{end} \rightarrow \{0, 1\}$, where $\mathcal{V}(s) = 1$ indicates a correct solution and $\mathcal{V}(s) = 0$ indicates an incorrect or incomplete solution. Then we partition the terminal nodes into correct and incorrect/incomplete subsets:
\begin{equation}
    \mathcal{S}_\text{correct}^{(k)} = \{s \in \mathcal{S}_\text{end}^{(k)} \mid \mathcal{V}(s) = 1\}, \quad 
    \mathcal{S}_\text{incorrect}^{(k)} = \{s \in \mathcal{S}_\text{end}^{(k)} \mid \mathcal{V}(s) = 0\}.
\end{equation}
If $\mathcal{S}_\text{correct}^{(k)} = \emptyset$, we employ an \emph{entropy-based selection} to identify the most confident wrong rollout. The terminal node with the lowest average entropy along its root-to-leaf trajectory is selected:
\begin{equation}
    s^*_\text{neg} = \arg\min_{s \in \mathcal{S}_\text{incorrect}^{(k)}} \bar{H}(\mathbf{t}(s)),
\end{equation}
where $\mathbf{t}(s) = (x, s_1, s_2, \ldots, s)$ represents the unique trajectory from root $x$ to terminal node $s$, and the average trajectory entropy is defined as $\bar{H}(\mathbf{t}(s)) = \frac{1}{|\mathbf{t}(s)|} \sum_{i=1}^{|\mathbf{t}(s)|} H(\pi_\theta(s_i \mid o_i)),$, where $H(\pi_\theta(s_i \mid o_i)) = -\sum_{a_{i,k}} \pi_\theta(a_{i,k} \mid o_i, a_{i,<k}) \log \pi_\theta(a_{i,k} \mid o_i, a_{i,<k})$ is the Monte Carlo estimation of the Shannon entropy of the token distribution at step $i$. $a_{i,k}$ is the $k$-th token of step $s_i$, and $a_{i,<k}$ denotes the tokens preceding $a_{i,k}$. This strategy prioritizes incorrect reasoning sequences with low decision uncertainty, targeting areas where the model is most confident in its decisions and would benefit from additional training and supervision. {We find that this most-confident incorrect selection consistently outperforms random and least-confident selection across all benchmarks, see Table \ref{tab:incorrect_selection}.}

\subsection{Heuristic Score Backup}

Let $\mathbf{t}^*$ denote the selected trajectory for backpropagation, which is either a correct solution trajectory or the most confident negative trajectory $\mathbf{t}(s^*_\text{neg})$ identified through entropy-based selection. Let $q^{(m)}(s_i)$ denote the \emph{q-value} for node $s_i \in \mathbf{t}^*$ after the $m$-th rollout backpropagation. We define the iterative q-value update rule for nodes along the selected trajectory:
\begin{equation}
q^{(m)}(s_i) = q^{(m-1)}(s_i) + \gamma(i,l) \cdot q^{(m)}(s_\text{end}),
\end{equation}
where $\gamma(i,l): \mathbb{Z}^+ \times \mathbb{Z}^+ \rightarrow [0,1]$ is the {\emph{depth decay}} function that assigns higher weights to nodes closer to the terminal node. It is defined as $\gamma(i,l) = \max\left(\frac{i}{l}, \gamma_{\min}\right)$, where $i$ is the current node index in the trajectory, $l$ is the terminal node index, and $\gamma_{\min} = 0.1$ is the minimum decay threshold.

The q-value initialization is $q^{(0)}(s_i) = 0$ for all $s_i \in \mathcal{T}$. Terminal node rewards are assigned according to the verification function's result:
\begin{equation}
\label{equ:terminal_score}
q(s_\text{end}) = \begin{cases}
+1 & \text{if } \mathcal{V}(s_\text{end}) = 1 \text{ (correct),} \\
-1 & \text{if } \mathcal{V}(s_\text{end}) = 0 \text{ (incorrect)} \vee d(s_\text{end}) < d_\mathcal{T} \text{ (incomplete).} \\
\end{cases}
\end{equation}

To ensure positive q-values (\emph{e.g.}, $q_{\text{correct}}=0.1$) for nodes on correct reasoning paths while penalizing nodes leading to incorrect or incomplete solutions, we enforce the constrained update rule:
\begin{equation}
\label{equ:backup}
    q^{(m)}(s_i) = \begin{cases}
    q^{(m-1)}(s_i) + \gamma(i,l) \cdot q^{(m)}(s_\text{end}) & \text{if } q^{(m-1)}(s_i) \cdot q^{(m)}(s_\text{end}) \geq 0, \\
    \gamma(i,l) \cdot q^{(m)}(s_\text{end})   & \text{ elif } q^{(m)}(s_\text{end}) > 0,\\    
     q^{(m-1)}(s_i) &  \text{ elif } q^{(m-1)}(s_i) > 0. \\    
\end{cases}
\end{equation}
This constraint preserves the invariant that $q^{(m)}(s_i) \geq 0$ for all intermediate nodes $s_i \in \mathcal{T} \setminus \mathcal{S}_\text{end}$ leading to correct solutions, while allowing negative values only for nodes observed on any correct trajectory under the current search process. {More justification is elucidated in Appendix~\ref{app:justify_backup}.}

\subsection{Hybrid Selection Strategy}

{UCT~\citep{kocsis2006bandit} is the standard selection rule used in classical MCTS to balance exploitation of high-value nodes and exploration of under-visited ones.}
Our MCTS employs a \emph{hybrid selection strategy} that combines traditional UCT-based local selection with novel global frontier selection, each serving distinct purposes in the search process.

\paragraph{Local Selection for Sibling Comparison}

During the expansion of a selected node, we generate multiple candidate children and must determine which to add to the tree. For this \emph{local sibling comparison}, we follow the traditional MCTS protocol and employ the UCT algorithm as$
\text{UCT}(s) = Q(s) + \lambda \sqrt{\frac{\ln N_{\text{parent}}(s)}{N(s)}}$, where $Q(s) = \frac{q(s)}{N(s)}$ represents the average reward per visit, $N_{\text{parent}}(s)$ is the number of visits from the parent node, and $\lambda$ balances exploitation and exploration. This local selection ensures that we make optimal decisions when choosing among sibling nodes that share the same parent and context.

\paragraph{Global Frontier Selection for Next Expansion}

After completing the first score backup phase, we need to identify the most promising node across the \emph{entire search tree} for the next expansion round. This is where our novel global frontier selection mechanism operates.

Unlike traditional MCTS, which performs root-to-leaf traversals using UCT at each level, our global approach directly compares all frontier nodes simultaneously. We maintain a global view of all leaf nodes across the entire search tree $\mathcal{T}$ and prioritize promising expansion points globally:
\begin{equation}
\label{equ:frontier_node}
    \mathcal{F} = \{s \in \mathcal{T} \mid \xi(s) = 0, s\notin \mathcal{S}_{\text{end}}, d(s) < d_{\mathcal{T}}\}. 
\end{equation}

For each frontier node $s \in \mathcal{F}$ and its associated observation (prefix) $o_s$, we compute a \emph{frontier priority score}:
\begin{equation}
\label{equ:frontier_score}
    F(s) = \underbrace{\lambda_1\times \tanh(Q_{\text{parent}}(s))}_{{\text{Quality Potential}}} +\underbrace{\lambda_2 \times H(\pi_\theta(s\mid o_s))}_{\text{Uncertainty Bonus}} + \underbrace{\lambda_3 \times D(d(s))}_{\text{Depth Bonus}}.
\end{equation}
Here, the quality potential term $\tanh(Q_{\text{parent}}(s))$ encourages the selection of nodes whose parents have demonstrated high value, using the tanh transformation to smoothly handle negative Q-values and map them to the range $[-1,1]$. 
The uncertainty bonus term $H(\pi_\theta(s\mid o_s))$ provides exploration guidance by adjusting priority according to the policy’s entropy; the sign of its coefficient can be utilized to steer selection toward regions with high confidence or uncertainty.
The depth bonus term $D(d(s))$ encourages deeper exploration by providing additional priority to nodes at greater depths, where we empirically find $D(d(s)) = \sqrt{d(s)/d_\mathcal{T}}$ to be most effective among other variants including $d(s)$ and $\log(d(s)+1)$. The node with the highest frontier score is selected for the next expansion: $s^* = \arg\max_{s \in \mathcal{F}} F(s)$.

\paragraph{Rationale for Hybrid Approach}

This hybrid design leverages complementary strengths: local UCT selection ensures principled sibling comparisons within subtrees, while global frontier selection mitigates UCT's myopia by allocating resources across subtrees. The approach achieves three key advantages: \emph{(1) Computational efficiency} by eliminating redundant root-to-leaf traversals, \emph{(2) Enhanced exploration coverage} by preventing the algorithm from getting trapped in locally promising but globally suboptimal subtrees, and 
\emph{(3) Uncertainty-guided search} that leverages the policy’s entropy to target regions expected to benefit from additional training supervision, with the bonus coefficient controlling the direction of this preference.

\section{Adaptive Training Strategy with Replay Buffer}

While MCTS offers fine-grained credit assignment, applying it to every training example is computationally infeasible. To address this, we adopt an iterative filtering strategy with a replay buffer mechanism that focuses MCTS computation on challenging examples while preventing catastrophic forgetting of solved problems. The complete pipeline is depicted in Algorithm~\ref{alg:DeepSearch}.

\subsection{Iterative Training with Progressive Filtering}

Our training process follows an iterative approach that progressively refines the training subset based on model performance. We begin by using the base RL model to perform an initial screening on the entire dataset $\mathcal{D}_{\text{hard}}$, creating the first training subset $\mathcal{D}_{\text{hard}}^{(0)}$ for MCTS-based RL training.

Specifically, the iterative training process proceeds as follows:

\textbf{Initial Subset Construction:} Given the base policy $\pi_{\theta^{(0)}}$, we evaluate its performance on the full training set $\mathcal{D}_{\text{train}}$ using direct rollouts and construct the initial hard subset:
\begin{equation}
\mathcal{D}_{\text{hard}}^{(0)} = \{ x \in \mathcal{D}_{\text{train}} \mid \texttt{Pass1@K}(x, \pi_{\theta^{(0)}}) < \delta^{(0)} \},
\end{equation}
where $\texttt{Pass1@K}(x, \pi)$ represents the success rate when sampling $K=4$ solutions for problem $x$ using policy $\pi$, and $\delta^{(0)}\in (0,1)$ is the initial filtering threshold.

\textbf{Iterative Refinement:} After each training phase $i$, we re-evaluate the updated policy $\pi_{\theta^{(i)}}$ on the current hard subset and apply threshold-based filtering to create the next iteration's training set:
\begin{equation}
\mathcal{D}_{\text{hard}}^{(i+1)} = \{ x \in \mathcal{D}_{\text{hard}}^{(i)} \mid \texttt{Pass1@K}(x, \pi_{\theta^{(i)}}) < \delta^{(i)} \}.
\end{equation}

The filtering threshold $\delta^{(i)}$ is typically set to 25\%, ensuring that only problems with insufficient success rates remain in the active training set. This progressive filtering concentrates computational resources on increasingly challenging problems as the model improves.

\subsection{Replay Buffer with Cached Solutions}

To prevent catastrophic forgetting and efficiently leverage previously discovered solutions, we maintain a replay buffer $\mathcal{R}$ that stores correct reasoning trajectories from earlier training phases.

\textbf{Buffer Population.} During each training iteration $i$, we identify problems that obtained correct solutions through MCTS rollouts but still fail to meet the filtering threshold after training:
\begin{equation}
    \mathcal{R}_{\text{candidates}}^{(i)} = \{ (x, \mathbf{t}_{\text{correct}}) \mid x \in \mathcal{D}_{\text{hard}}^{(i)}, \exists \mathbf{t}_{\text{correct}} \in \mathbb{T}(x), \texttt{Pass1@K}(x, \pi_{\theta^{(i)}}) < \delta^{(i)} \}.
\end{equation}
These candidate trajectories are added to the replay buffer, attaining $\mathcal{R}^{(i+1)} = \mathcal{R}^{(i)} \cup \mathcal{R}_{\text{candidates}}^{(i)}.$

\textbf{Cached Solution Usage.} Instead of randomly sampling from the replay buffer, we employ a deterministic strategy that directly utilizes cached solutions when available. For each problem $x$ in the current training iteration, we first check whether a correct solution has been previously cached. This approach eliminates redundant MCTS computation for problems with known solutions while directing computational resources toward truly challenging, unsolved problems.

\textbf{Hybrid Rollout Strategy.} When processing problems in the current hard subset $\mathcal{D}_{\text{hard}}^{(i)}$, we apply different rollout strategies based on cache availability:
\begin{equation}
    \text{Rollout}(x) = \begin{cases}
    \mathbf{t}_{\text{cached}} \cup \text{DirectRollouts}(x, \beta) & \text{if } (x, \mathbf{t}_{\text{cached}}) \in \mathcal{R}^{(i)}, \\
    \text{MCTS}_{\text{full}}(x) & \text{otherwise.}
    \end{cases}
\end{equation}
For problems with cached solutions, we directly incorporate the stored correct trajectory $\mathbf{t}_{\text{cached}}$ and supplement it with $\text{DirectRollouts}(x, \beta)$, which samples $\beta \cdot B$ additional solution attempts from the current policy $\pi_\theta(\cdot|x)$, where $\beta\in [0,1]$ {is implicitly determined by the number of cached solutions per problem (we allocate fewer DirectRollouts as more correct trajectories are cached)} and $B$ is the standard sampling budget. For problems without cached solutions, we apply the complete MCTS search process $\text{MCTS}_{\text{full}}(x)$. Moreover, among the incorrect samples, we remove data containing garbled text or infinite repetitions. Based on empirical evidence, optimizing policies on such problematic data frequently leads to training collapse~\citep{bai2025intern}. The training dataset for each iteration is then constructed as:
\begin{equation}
\begin{split}
    \mathcal{T}_{\text{train}}^{(i)} 
    = \underbrace{\bigcup_{x: (x,\mathbf{t}_{\text{cached}}) \in \mathcal{R}^{(i)}} \left\{ \mathbf{t}_{\text{cached}} \cup \text{DirectRollouts}(x, \beta) \right\}}_{\text{Cached problems}} \cup \underbrace{\bigcup_{x: (x,\mathbf{t}_{\text{cached}}) \notin \mathcal{R}^{(i)}} \text{MCTS}_{\text{full}}(x)}_{\text{Unsolved problems}}.
\end{split}
\end{equation}
This eliminates the need for artificial sampling ratios or complex batch composition strategies, as training data naturally incorporates both preserved knowledge and fresh exploration, tailored to problem-specific requirements. This has three key benefits: \emph{(1) Computational efficiency} by avoiding redundant MCTS computation, \emph{(2) Solution preservation} by guaranteeing the inclusion of cached correct trajectories, and \emph{(3) Continued exploration} at minimal computational cost.

\subsection{Tree-GRPO Training Objective}

After constructing a search tree $\mathcal{T}$ for a sample question $x$ in the dataset $\mathcal{D}_{\text{train}}$, we develop our Tree-GRPO training objective. This objective combines q-value regularization with policy optimization to learn effectively from tree-structured reasoning traces.

\paragraph{Q-Value Soft Clipping.} To address the q-value explosion problem for intermediate nodes while preserving meaningful gradients, we first apply \emph{soft clipping} using the hyperbolic tangent function:
\begin{equation}
\label{equ:clip}
    q(s_j) = \tanh\left({q^{(k_{\max})}(s_j)}/{\epsilon_q}\right) \cdot q_{\max} \quad \text{for all } s_j \in \mathcal{T} \setminus \mathcal{S}_\text{end}
\end{equation}
where $k_{\max}$ is the maximum rollout iterations, $\epsilon_q=1.0$ is the temperature parameter, and $q_{\max}=1$ defines the maximum allowable q-value magnitude. 

This soft clipping approach prevents q-value explosion by maintaining all intermediate node q-values within $[-q_{\max}, q_{\max}]$, while offering several key advantages: \emph{(i)} it naturally bounds q-values without hard discontinuities, \emph{(ii)} it preserves gradients everywhere, preventing the zero-gradient problem that occurs with hard clipping when all values hit the same bound, and \emph{(iii)} it maintains the relative ordering of q-values while compressing extreme outliers. Terminal node q-values remain unchanged as defined in Eq.~\ref{equ:terminal_score}. {In practice, fewer than about 5\% of intermediate node q-values fall into the saturation region and less than 0.5\% lie near the boundaries, so clipping mainly compresses pathological tails rather than altering the bulk of the distribution.}

\paragraph{Training Objective.} With regularized q-values, we formulate and maximize the Tree-GRPO objective as:
\begin{equation}
\begin{split}
\label{equ:tree_loss}
    \mathcal{J}(\theta)=\mathbb{E}_{\mathbb{T} \sim \mathcal{T}, \mathbf{t}^i\sim\mathbb{T}, (s_j, o_j) \sim \mathbf{t}^i}
    \frac{1}{\left|s_j\right|} \sum_{k=1}^{\left|s_j\right|} \min \left(\rho_{j,k}(\theta) \hat{A}_{j,k}, \operatorname{clip}\left(\rho_{j,k}(\theta), 1-\epsilon_{\mathrm{low}}, 1+\epsilon_{\mathrm{high}}\right) \hat{A}_{j,k}\right) 
\end{split}
\end{equation}
where $\rho_{j,k}(\theta)=\frac{\pi_\theta\left(a_{j,k} \mid o_j, a_{j,<k}\right)}{\pi_{\theta_{\text {old }}}\left(a_{j,k} \mid o_j, a_{j, <k}\right)}$ is the importance ratio. The parameters $\epsilon_{\text {high}}$ and $\epsilon_{\text {low}}$ follow the Clip-Higher strategy of DAPO~\citep{yu2025dapo}, while we also remove the KL regularization term $\mathbb{D}_{\mathrm{KL}}$ to naturally diverge~\citep{luo2025deepcoder,he2025skywork}. 
An overlong buffer penalty is applied to responses that exceed a predefined maximum of 4096. The advantage function for node $s_j$ in trajectory $\mathbf{t}_i$ is computed using \emph{sequence-level normalization}~\citep{chu2025gpg}: 
\begin{equation}
\label{equ:adv_norm}
    \hat{A}_{j,k} = q(s_j) - \mu_{\mathbf{t}}, 
\end{equation}
where $\mu_{\mathbf{t}}$ is the average reward of the terminal nodes $\mathcal{S}_\text{end}$ throughout the tree $\mathbb{T}$.  This normalization is crucial in practice, particularly for mitigating uncontrolled growth in response length.
Notably, Tree-GRPO can be degraded to the vanilla DAPO if we consistently leverage the outcome reward $q(s_\text{end})$ as $q(s_j)$ for all intermediate nodes. 

\section{Experiments}

\subsection{Benchmark Performance Evaluation}

\paragraph{Datasets and Base Models.} We train DeepSearch based on {Nemotron-Research-Reasoning-Qwen-1.5B v2}~\citep{liu2025prorl} and employ DeepMath-103K~\citep{he2025deepmath} as the raw dataset. DeepMath-103K is a large-scale mathematical dataset designed to be highly challenging and rigorously decontaminated across numerous benchmarks.
We evaluate DeepSearch against state-of-the-art 1.5B reasoning models on six mathematical benchmarks: AIME 2024/2025, AMC2023, MATH500~\citep{hendrycks2021measuring}, Minerva~\citep{lewkowycz2022solving}, and Olympiad~\citep{he2024olympiadbench}. More experimental details are described in Appendix~\ref{app:details}.

\paragraph{Baselines.} We compare against recent 1.5B models spanning different mechanisms: base models (Qwen2.5-Math variants), RL-trained models (DeepSeek-R1-Distill, STILL-3~\citep{Slow_Thinking_with_LLMs_3_Preview}, Open-RS series~\citep{dang2025reinforcement}, advanced RL methods (DeepScaleR~\citep{deepscaler2025}, Nemotron variants), and search-based approaches (Qwen2.5-Math-Oat-Zero~\citep{liu2025understanding}). Our evaluation methods and results are consistent with~\citet{hochlehnert2025sober}.

\paragraph{Results.} Table~\ref{tab:math_results} shows DeepSearch-1.5B achieves 62.95\% average accuracy, outperforming all baselines, including the previous best Nemotron-Research-Reasoning-Qwen-1.5B v2 (61.70\%). DeepSearch-1.5B demonstrates consistent improvements across all benchmarks, with notable gains on AIME 2024 (53.65\% vs 51.77\%) and AMC (90.39\% vs 88.83\%). The 1.25 percentage-point improvement over the previous state of the art validates the effectiveness of integrating structured search into RLVR training rather than restricting it to inference-only use.
\begin{table*}[t]
\centering
\caption{Performance comparison of 1.5B-scale language models on standard mathematical reasoning benchmarks. We report \texttt{Pass1@1} accuracy estimated with $n=32$ samples. The best-performing results are highlighted in bold. All evaluations were conducted on a 128$\times$H100 96G cluster.}
\label{tab:math_results}
\resizebox{\textwidth}{!}{%
\begin{tabular}{l|cccccc|c}
\toprule
\textbf{Model} & \textbf{AIME24} & \textbf{AIME25} & \textbf{AMC23} & \textbf{MATH} & \textbf{Minerva} & \textbf{Olympiad} & \textbf{Avg} \\
\midrule
Qwen2.5-Math-1.5B & 8.33 & 6.35 & 44.06 & 66.67 & 18.42 & 30.74 & 29.10 \\
Qwen2.5-Math-1.5B-Instruct & 10.10 & 8.85 & 55.08 & 74.83 & 29.32 & 40.00 & 36.37 \\
DeepSeek-R1-Distill-Qwen-1.5B & 31.15 & 24.06 & 72.81 & 85.01 & 32.18 & 51.55 & 49.46 \\
STILL-3-1.5B & 31.46 & 25.00 & 75.08 & 86.24 & 32.77 & 53.84 & 50.73 \\
Qwen2.5-Math-1.5B-Oat-Zero & 20.00 & 10.00 & 52.50 & 74.20 & 26.84 & 37.78 & 36.89 \\
Open-RS1-1.5B & 30.94 & 22.60 & 73.05 & 84.90 & 29.92 & 52.82 & 49.04 \\
Open-RS2-1.5B & 28.96 & 24.37 & 73.52 & 85.06 & 29.74 & 52.63 & 49.05 \\
Open-RS3-1.5B & 30.94 & 24.79 & 72.50 & 84.47 & 29.11 & 52.25 & 49.01 \\
DeepScaleR-1.5B & 38.54 & 30.52 & 80.86 & 88.79 & 36.19 & 58.95 & 55.64 \\
Nemotron-Research-Reasoning-Qwen-1.5B v1 & 45.62 & 33.85 & 85.70 & 92.01 & 39.27 & 64.56 & 60.17 \\ 
Nemotron-Research-Reasoning-Qwen-1.5B v2 & 51.77 & 32.92 & 88.83 & 92.24 & 39.75 & 64.69 & 61.70 \\ \midrule 
DeepSearch-1.5B & \textbf{53.65}  & \textbf{35.42} & \textbf{90.39} & \textbf{92.53} & \textbf{40.00} & \textbf{65.72} & \textbf{62.95} \\ \bottomrule
\end{tabular}}
\vspace{-2em}
\end{table*}

\subsection{Training Efficiency Analysis}

To evaluate the practical viability of DeepSearch, we compare computational costs with extended training approaches that scale solely with the number of additional training steps. As shown in Table~\ref{tab:efficiency}, extended training exhibits diminishing returns: 325 additional steps achieve 61.78\% accuracy using 326.4 GPU hours, whereas 1,875 steps plateau at 62.02\% accuracy despite consuming 1,883.2 GPU hours. This reveals the fundamental limitation of depth-first scaling, where performance gains become marginal as computational investment grows exponentially.

DeepSearch achieves superior results through algorithmic innovation rather than brute-force computation. With only 50 additional training steps, DeepSearch reaches 62.95\% accuracy using 330 GPU hours—outperforming the most extensive baseline (1,883.2 hours) while using 5.7$\times$ fewer resources. This efficiency stems from a structured search that extracts maximum value from each training step by systematically exploring diverse solution paths.

\begin{wrapfigure}{r}{0.5\textwidth}
    \centering
    \vspace{-1.5em}
    \includegraphics[width=0.50\textwidth]{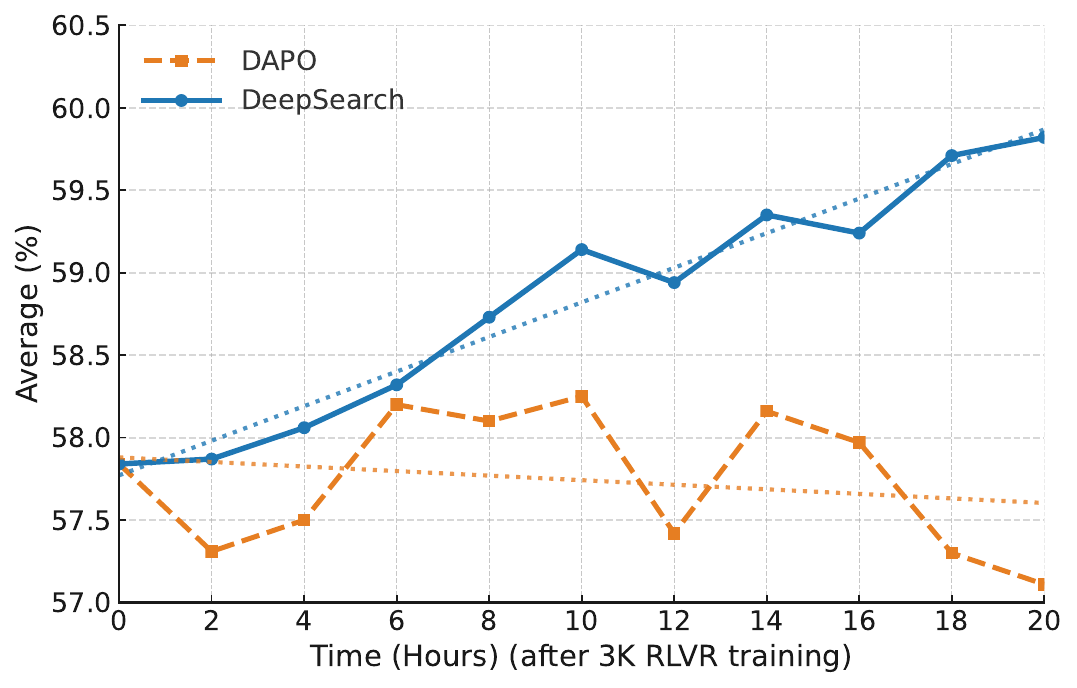}
    \vspace{-2em}
    \caption{ Average performance ({AIME 2024}, {AIME 2025}, and {AMC 2023}) of \textbf{DAPO} and \textbf{DeepSearch} after 3K RLVR training.  Markers denote evaluations, while dotted lines indicate linear trends.}
    \vspace{-0.8em}
    \label{fig:training_curve}
\end{wrapfigure}
Figure~\ref{fig:training_curve} illustrates the training dynamics over 20 hours following 3K RLVR training. DAPO exhibits gradual linear improvement with a shallow slope, while DeepSearch demonstrates more efficient learning through structured exploration. The superior convergence properties suggest that RLVR bottlenecks stem from the quality of exploration rather than from insufficient training time.

These results challenge the common assumption that improving RLVR performance requires proportional increases in computational budget. Compared with the training of Nemotron-Research-Reasoning-Qwen-1.5B v2, DeepSearch-1.5B achieves a 72× efficiency improvement, highlighting a qualitative shift toward algorithmically driven optimization rather than resource-driven scaling. This suggests that systematic exploration can offer a more practical and scalable path for advancing RLVR capabilities than extended high-compute training.
\begin{table}[t]
\centering
\caption{Comparison of methods on efficiency and performance, which are trained from DeepSeek-R1-Distill-Qwen-1.5B.}
\resizebox{\textwidth}{!}{
\begin{tabular}{l|l|ccccc}
\toprule
\textbf{Method} & \textbf{RLVR} & \textbf{Steps} & \textbf{Samples (K)} & \textbf{Time (h)} & \textbf{GPU Hours} & \textbf{Math Score} \\
\midrule
DeepSeek-R1-Distill-Qwen-1.5B  & --   & --  &   --   &   --    &   --   & 49.46 \\
Nemotron-Research-Reasoning-Qwen-1.5B v1 & DAPO  & 2000 &  -- & -- & 16000 & 60.10 \\ 
Nemotron-Research-Reasoning-Qwen-1.5B v2 & DAPO   & 3000  &   --   &   --  & 24000   & 61.70 \\  \midrule 
Extended Training  & DAPO & +325 & 665.6 & 20.4 & 326.4 & 61.78 \\
Extended Training  & DAPO + KL & +785 & 1607.7  & 49.3  & 788.8  & 62.08 \\
Extended Training  & DAPO + KL & +1875 & 3840.0 & 117.7  & 1883.2  & 62.02 \\ \midrule 
DeepSearch-1.5B  & Tree-GRPO  & +50   & 102.4  & 20.6   & 330   & \textbf{62.95} \\ \bottomrule
\end{tabular}} 
\vspace{-1.5em}
\label{tab:efficiency}
\end{table}


\subsection{Search Strategy Ablation}

Table~\ref{tab:search_ablation} compares our global frontier selection against vanilla UCT under different configurations on 1.2K samples from extremely hard DeepMath-103K problems.

\textbf{Global vs. Local Selection.} Our global frontier selection ($\lambda_1=0.4$) reduces iterations by $10.4\%$ ($209.6 \rightarrow 187.7$) and improves trajectory rewards ($-0.82 \rightarrow-0.65)$ compared to vanilla UCT, while maintaining similar search depth and entropy. This demonstrates that comparing frontier nodes across the entire tree is more efficient than traditional root-to-leaf UCT traversals.

\textbf{Depth Bonus Impact.} We evaluate three depth bonus functions $D(d(s))$: (i) Logarithmic $\log (d(s)+1)$ provides minimal improvements, (ii) Linear $d(s)$ achieves the most aggressive efficiency gains with 59\% reduction in per-tree time ($1179.6$s $\rightarrow 480.9$s) and deepest exploration (21.55 depth), but at cost of solution quality ( -0.76 reward), (iii) Square root $\sqrt{d(s) / d_T}$ offers the best balance, maintaining search quality (-0.65 reward) with significant computational savings.

\textbf{Uncertainty Bonus.} Adding uncertainty weighting ($\lambda_2=0.4$) increases exploration diversity (entropy $1.23 \rightarrow 1.31$) by prioritizing high-uncertainty policy regions, but introduces computational variability ($92.5 \pm 22.5$ iterations).

\textbf{Configuration Selection.} We adopt $\sqrt{d(s) / d_T}$ with $\lambda_1=0.4, \lambda_3=0.01$ as our default, balancing computational efficiency (189.3 iterations), search quality (-0.65 reward), and stable performance. This configuration eliminates UCT's redundant traversals while maintaining principled exploration through quality potential and depth guidance.

\begin{table}[t]
    \centering
    \caption{Ablation study of different search strategies in DeepSearch. We compare vanilla UCT with our proposed global frontier selection under varying depth bonus functions $D(d(s))$. Reported metrics include search statistics such as average search depth, trajectory entropy, and trajectory reward, as well as computational cost, measured by the number of iterations, average per-iteration time (in seconds), and per-tree time (in seconds). Results are presented as mean $\pm$ standard deviation.}
    \resizebox{\textwidth}{!}{%
    \begin{tabular}{lcccccccc}\toprule
     \multirow{2}{*}{\textbf{Method}}  & \multirow{2}{*}{$D(d(s))$} & \multicolumn{3}{c}{\textbf{Search Metrics}} &  \multicolumn{3}{c}{\textbf{Computational Cost}}  \\ \cmidrule(lr){3-5} \cmidrule(lr){6-8}
     & & Depth  & Entropy & Reward & Num. Iter. & Time Per Iter.  & Time Per Tree\\ \midrule  
     Vanilla UCT & -- & $20.11 \pm 4.72$ & $1.23 \pm 0.29$ & $-0.82 \pm 0.57$ & $209.6 \pm 14.8$ & $5.63 \pm 0.21$ & $1179.6 \pm 95.0$ \\ 
     \midrule
     \multicolumn{8}{l}{\textbf{Global Frontier Selection}} \\
     $\lambda_1 = 0.4$ & -- & $20.28 \pm 4.80$ & $1.23 \pm 0.29$ & $-0.65 \pm 0.76$ & $187.7 \pm 16.2$ & $5.76 \pm 0.19$ & $1087.7 \pm 105.0$ \\ 
     $\lambda_1 = 0.4,\lambda_3=0.01$ & $\log(d(s)+1)$ & $20.33 \pm 4.77$ & $1.23 \pm 0.30$ & $-0.65 \pm 0.76$ & $185.5 \pm 15.9$ & $5.85 \pm 0.19$ & $1080.3 \pm 102.2$ \\
     $\lambda_1 = 0.4,\lambda_3=0.01$ & $d(s)$ & $21.55 \pm 5.13$ & $1.24 \pm 0.29$ & $-0.76 \pm 0.65$ & $85.7 \pm 7.7$ & $5.61 \pm 0.12$ & $480.9 \pm 41.9$ \\ 
     $\lambda_1 = 0.4,\lambda_2=0.4,\lambda_3=0.01$ & $\sqrt{d(s)/d_\mathcal{T}}$ & $20.83 \pm 4.71$ & $1.31 \pm 0.30$ & $-0.79 \pm 0.62$ & $92.5 \pm 22.5$ & $5.48 \pm 0.13$ & $505.2 \pm 114.8$ \\ 
     \rowcolor{black!05} $\boldsymbol{\lambda_1 = 0.4,\lambda_3=0.01}$ & $\boldsymbol{\sqrt{d(s)/d_\mathcal{T}}}$ & $\boldsymbol{20.29 \pm 4.83}$ & $\boldsymbol{1.24 \pm 0.29}$ & $\boldsymbol{-0.65 \pm 0.76}$ & $\boldsymbol{189.3 \pm 14.7}$ & $\boldsymbol{5.66 \pm 0.14}$ & $\boldsymbol{1070.7 \pm 87.3}$ \\ 
     \bottomrule
    \end{tabular}}
    \vspace{-1.5em}
    \label{tab:search_ablation}
\end{table}

\subsection{Algorithm Evolution and Component Contributions} 

To understand the individual contributions of each component, we present a systematic ablation study of our DeepSearch algorithm in Table~\ref{tab:evolution}. Starting from the \texttt{Nemotron-Qwen-1.5B v2} base, we incrementally add components and analyze their impact:

(i) \textbf{Vanilla DeepSearch Foundation.}  We begin with a basic MCTS integration using a simple q-value update rule: 
\[ q^{(m)}(s_i) =
\begin{cases}
q^{(m-1)}(s_i) + \gamma(i,l)\cdot q^{(m)}(s_\text{end}) & \text{if } q^{(m-1)}(s_i)\cdot q^{(m)}(s_\text{end}) \ge 0,\\
\max\!\left(q^{(m-1)}(s_i) + \gamma(i,l)\cdot q^{(m)}(s_\text{end}),\, 0\right) & \text{otherwise.}
\end{cases}
\]
This assigns constant values to nodes along correct reasoning paths but yields only limited improvement over the baseline.  
(ii) \textbf{Enhanced Q-Value Updates with Outcome Rewards.} We replace the simple update with our constrained backup rule (Eq.~\ref{equ:backup})  and use outcome-based advantages $\hat{A}_{j,k} = q(s_\text{end})$ for all nodes. This provides more stable credit assignment and yields meaningful improvements.
(iii) \textbf{Fine-Grained Node-Level Advantages.} Moving beyond outcome-only rewards, we assign node-specific advantages $\hat{A}_{j,k} = q(s_j)$ based on each node's individual q-value. This enables more precise credit assignment across different reasoning steps.
(iv) \textbf{Standard Advantage Normalization.}  We implement standard normalization as $\hat{A}_{j,k} = \frac{q(s_j) - \mu_{\mathbf{t}}}{\sigma_{\mathbf{t}} + \varepsilon},$ where $\sigma_{\mathbf{t}}$ is the standard deviation of the rewards of the terminal nodes $\mathcal{S}_\text{end}$ throughout the tree $\mathbb{T}$.  The constant $\varepsilon$ prevents numerical instability when the variance is small. This stabilizes training but introduces variance-based scaling.
(v) \textbf{Mean-Only Normalization.} We adopt mean-only normalization (Eq.~\ref{equ:adv_norm}). This addresses miscalibration issues in GRPO while maintaining stable advantage scaling.~\citep{bereket2025uncalibrated}.
(vi) \textbf{Global Frontier Selection.} Finally, we integrate our novel frontier selection strategy (Eq.~\ref{equ:frontier_node}), which prioritizes promising expansion candidates across the entire search tree rather than following traditional root-to-leaf UCT-like traversals.

{Moreover, to justify our choice of prioritizing the \emph{most confident} incorrect trajectory when no correct solution is found,  we compare three selection strategies in Tab.~\ref{tab:incorrect_selection}. 
Selecting the trajectory with the lowest average entropy, representing the model’s \emph{most confident} mistake, achieves the best average performance (+0.86 over random and +1.05 over least confident),  and is the only method that consistently improves results across all benchmarks. 
This confirms that high-confidence incorrect trajectories expose \emph{systematic reasoning errors} that benefit most from targeted supervision, whereas random or low-confidence errors tend to be noisy or uninformative.} 

The results demonstrate that each component contributes meaningfully to the final performance, with frontier selection providing the largest single improvement. The cumulative effect shows that systematic exploration and fine-grained credit assignment are essential for maximizing the benefits of search-augmented RLVR.

\vspace{-3mm}
\begin{table}[t]
\centering
\caption{Ablation study illustrating the step-by-step evolution of \textbf{DeepSearch}. Starting from Vanilla DeepSearch with a simple $q$-update, we progressively add outcome-reward–based and fine-grained advantages, standard or mean-only normalization, and frontier node selection.}
\resizebox{\textwidth}{!}{%
\begin{tabular}{@{}l c c c c c c c@{}} \toprule
\textbf{Model / Change} & 
\textbf{AIME24} & \textbf{AIME25} & \textbf{AMC23} & \textbf{MATH} & \textbf{Minerva} & \textbf{Olympiad} & \textbf{Avg} \\ \midrule
Nemotron-Research-Reasoning-Qwen-1.5B v2 & 51.77 & 32.92 & 88.83 & 92.24 & 39.75 & 64.69 & 61.70 \\ \addlinespace[4pt]
\rowcolor{black!05} \;+\;Vanilla DeepSearch  & 51.98 & 34.06 & 86.64 & 87.00 & 37.96 & 64.00 & 60.27 \\ 
\rowcolor{black!05} \;+\;New $q$ Update \& Coarse-grained Token Scores & 51.04 & \textbf{35.73} & 86.48 & 90.66 & 39.14 & 65.23 & 61.38 \\ 
\rowcolor{black!05} \;+\;New $q$ Update \& Fine-grained Token Scores & 50.52 & 35.52 & 88.83 & 91.70 & 39.71 & 64.81 & 61.85 \\
\rowcolor{black!05} \;+\;Standard Advantages Normalization & 52.60 & 35.00 & 89.30 & 92.44 & 39.29 & 64.99 & 62.27 \\
\rowcolor{black!05} \;+\;Mean-only Advantages Normalization & 51.98 & \textbf{35.73} & 89.06 & 91.88 & 39.58 & 65.71 & 62.32 \\
\rowcolor{black!05} \;+\;Frontier Selection  & \textbf{53.65} & 35.42 & \textbf{90.39} & \textbf{92.53} & \textbf{40.00} & \textbf{65.72} & \textbf{62.95} \\ \bottomrule
\end{tabular}}
\vspace{-1.5em}
\label{tab:evolution}
\end{table}

\begin{table}[t]
\centering
\caption{\textbf{Ablation on different trajectory selection strategies.} 
We compare using the most confident incorrect trajectory, a randomly selected incorrect trajectory, and the least confident incorrect trajectory. Confidence is measured via average trajectory entropy.}
\resizebox{\textwidth}{!}{%
\begin{tabular}{l|ccccccc} \toprule
\textbf{Method} & \textbf{AIME24} & \textbf{AIME25} & \textbf{AMC23} & \textbf{MATH} & \textbf{Minerva} & \textbf{Olympiad} & \textbf{Avg} \\ \midrule
Nemotron-Research-Reasoning-Qwen-1.5B v2 & 51.77 & 32.92 & 88.83 & 92.24 & 39.75 & 64.69 & 61.70 \\ \addlinespace[4pt]
\rowcolor{black!05} \;+\;Random incorrect trajectory & 52.41 & 34.23 & 89.31 & 92.21 & 39.97 & 64.41 & 62.09 \\
\rowcolor{black!05} \;+\;Least confident incorrect trajectory & 52.16 & 33.77 & 88.68 & 91.96 & 40.11 & 64.73 & 61.90 \\ 
\rowcolor{black!05} \;+\;\textbf{Most confident incorrect trajectory} & \textbf{53.65} & \textbf{35.42} & \textbf{90.39} & \textbf{92.53} & \textbf{40.00} & \textbf{65.72} & \textbf{62.95} \\ \bottomrule
\end{tabular}}
\vspace{-1.5em}
\label{tab:incorrect_selection}
\end{table}

\subsection{Discussion: Why Tree-Based Exploration Helps}

{Direct rollouts from $\pi_\theta$ behave like blind sampling: they quickly collapse into high-probability but low-diversity regions, rarely reaching deeper reasoning paths. As a result, they fail to uncover the informative trajectories required to correct systematic reasoning errors. In contrast, MCTS provides \emph{structured} exploration. Instead of distributing compute uniformly at random, it allocates search effort according to frontier quality, uncertainty, and depth. This yields three intuitive benefits:}

{(1) \textbf{Promising branches are expanded first.}   Frontier scoring prioritizes nodes with high value potential, preventing premature collapse into shallow modes.}
({2) \textbf{Overconfident mistakes become learnable.}   When no correct solution exists, we select the lowest-entropy incorrect trajectory, exposing systematic reasoning failures that direct rollouts rarely sample.}
{(3) \textbf{Deep reasoning chains are reachable.}  Depth bonuses encourage exploration of multi-step trajectories that direct rollouts fail to reach due to early-token drift.}
\vspace{-1em}

\section{Conclusion}

We introduced DeepSearch to integrate tree-based search into RLVR training, thereby addressing exploration bottlenecks that lead to performance plateaus. 
It achieves an average accuracy of 62.95\% on mathematical reasoning benchmarks, establishing a new state-of-the-art for 1.5B models while using 5.7× fewer GPU hours. This suggests that systematic exploration during training can, in many cases, provide greater gains than simply extending training time, highlighting the value of scaling training breadth through algorithmic innovation rather than relying solely on deeper compute.



\section*{Acknowledgment}
This work was supported in part by RS-2024-00457882, National AI Research Lab Project.


\bibliography{cite}
\bibliographystyle{iclr2026_conference}

\appendix
\newpage
\section{Experimental Details}
\label{app:details}

This section provides comprehensive details of our experimental setup, including system implementation, training configurations, MCTS parameters, optimization strategies, and evaluation protocols used in our DeepSearch framework.

\subsection{Training Data and Configuration}

We implement our DeepSearch system using the veRL framework \citep{sheng2024hybridflow}. Training is conducted on 16×H100 GPUs, each with 96GB of memory; evaluation is run on a larger 128×H100 cluster to reduce wall-clock time. The policy model is initialized with Nemotron-Research-Reasoning-Qwen-1.5B v2~\citep{liu2025prorl} (updated July 23rd). To ensure a fair comparison with a well-aligned policy, we additionally perform DAPO-based extended training using the Nemotron-Research-Reasoning-Qwen-1.5B-v2 initialization, with the same training configuration as DeepSearch.

Our training methodology uses the DeepMath-103K~\citep{he2025deepmath} dataset as $D_{\text{train}}$ and employs a DeepScaleR-style prompt template that instructs the model to ``Let's think step by step and output the final answer within \textbackslash{}boxed\{\}.'' To manage computational constraints, we truncate the left side, keeping the most recent tokens with a maximum prompt length of 2,048 tokens and limiting response generation to 16,384 tokens. The training process utilizes a global batch size of 256 samples, implemented through the DAPO-style Dynamic Batching strategy~\citep{yu2025dapo} to optimize memory utilization and training efficiency.

\subsection{Monte Carlo Tree Search Implementation}

Our MCTS implementation incorporates several strategic design choices to balance search efficiency and solution quality. The exploration coefficient ($\lambda$) for UCT Local Selection is set to 2.0, providing an optimal exploration-exploitation trade-off for mathematical reasoning tasks. The search architecture operates with a maximum depth of 64 levels, with each node allocated 256 tokens and expanding 8 children during the expansion phase. 
For entropy-based selection, we estimate average trajectory entropy using only tokens that appear in the response, rather than the entire per-position vocabulary, for computational efficiency. This average token negative log-probability along the sampled trajectory serves as a proxy for decision confidence.

To enhance search effectiveness, the system employs Global Frontier Selection for backtracking and applies a square-root function to depth-based bonuses, encouraging deeper exploration when beneficial. The global $\lambda_3$ parameter is configured to 0.01 for our frontier priority scoring, while an overlong buffer of 4,096 tokens with a penalty factor of 1.0 accommodates lengthy reasoning chains typical in complex mathematical problems.

\subsection{Advantage Estimation and Optimization}

For advantage estimation, we implement the Grouped Relative Policy Optimization (GRPO)~\citep{zhihong2024deepseekmath} estimator with sibling mean normalization to ensure stable learning dynamics. The Q-value soft clipping mechanism operates at a temperature of 1.0 with the maximum q-value magnitude set to 1.0, while incomplete trajectories receive a penalty score of $-1.0$ to discourage premature termination. Standard deviation normalization is disabled to prevent numerical instability during training.

The actor model is optimized using AdamW with a conservative learning rate of $1 \times 10^{-6}$ and 10 warmup steps, along with weight decay of 0.1 and gradient clipping at 1.0 to stabilize convergence. We follow the Clip-Higher strategy in DAPO~\citep{yu2025dapo} and set the lower and upper clipping thresholds to 0.2 and 0.28, respectively, with a clipping ratio of 10.0. Training proceeds with mini-batches of 32 samples per policy update using token-mean loss aggregation, while dynamic batch sizing accommodates up to 18,432 tokens per GPU. The entropy coefficient is set to 0 for pure exploitation, and the KL divergence loss is disabled to maximize performance on the target mathematical reasoning tasks.

\subsection{Sampling and Reward Configuration}

During rollout generation, we set the sampling temperature to 1.0 and $top\_p$ to 1.0, and disable $top\_k$ filtering to maintain diverse response generation. The system generates 8 rollouts per prompt, aligning with the expansion width parameter, within a context length of 18,432 tokens. This configuration ensures comprehensive exploration of the solution space while maintaining computational feasibility.
During evaluation, we uniformly use a low temperature of 0.6 and $top\_p$ of 0.95.

Our reward system implements a custom mathematical scoring function based on (\texttt{compute\_score}) from the \texttt{math\_dapo.py} module, designed to evaluate mathematical reasoning accuracy. We extract the final boxed answer by locating the last occurrence of \textbackslash{}boxed\{\} in the trajectory and apply the same text-normalization logic as veRL’s DAPO recipe to both prediction and ground-truth. The reward mechanism handles responses up to 16,384 tokens, following ProRL~\citep{liu2025prorl} and ensuring consistent evaluation across varying response lengths.

\subsection{Training Protocol}

The complete training protocol spans 100 steps with model checkpointing performed every 5 steps. This frequent checkpointing strategy ensures robust model preservation and enables detailed analysis of learning progress throughout training.

\section{Justification of DeepSearch Design Choice}
\subsection{Maximum Depth and Expansion Length}

\paragraph{{Data-driven determination of maximum search depth.}} We set the maximum search depth through two jointly defined factors: a rollout budget of 16{,}384 tokens and an expansion length of 256 tokens per node, yielding an approximate upper bound of $16{,}384 / 256 \approx 64$ expansion steps. In practice, many trajectories terminate earlier, so this acts as an upper bound rather than a hard requirement. Most importantly, this limit is derived from empirical statistics of the reasoning-length distribution in DeepMath-103K, which confirms that 16{,}384 tokens comfortably cover full reasoning chains for nearly all problems. The cap, therefore, reflects a principled trade-off between modeling requirements and training efficiency—not an arbitrary constant.

\paragraph{{Empirical validation of 256-token expansion length.}}
{
We evaluate several expansion lengths (128, 256, 512, 1024 tokens) and observe a clear trade-off:
\begin{itemize}
\item \textbf{Search-space exploration:} Longer expansions reduce tree breadth and make MCTS less likely to discover correct solution paths.
\item \textbf{Computational cost:} Shorter expansions significantly increase prefix re-encoding (“prefill”) costs, which dominate GPU time in MCTS exploration.
\end{itemize}
The 256-token step size consistently gives the best balance between coverage, training stability, and computational efficiency across all experiments.}

\paragraph{{Summary.}}
{
Both design choices of the depth cap and the expansion length arise from data-driven analysis and controlled comparisons. Together, they ensure that DeepSearch maintains wide search coverage while avoiding unnecessary computational overhead.
}

\subsection{Rationale Behind the Asymmetric Q-Value Backup Rule}
\label{app:justify_backup}

{Our constrained backup rule in Eq.~\ref{equ:backup} maintains for every intermediate node a scalar q-value representing its estimated tendency to lead toward a correct solution under the current policy and search process. When a terminal trajectory returns a reward signal ($+1$ for correct, $-1$ for incorrect or incomplete), the update is \emph{asymmetric}, depending on the relationship between the node’s current q-value and the reward.}

\paragraph{{Same-sign reinforcement.}}
{
When a node’s existing q-value and the terminal reward share the same sign, the update simply adds the depth-decayed reward. Thus, nodes that frequently appear on correct trajectories naturally accumulate stronger positive evidence, while nodes that are consistently involved in incorrect trajectories accumulate stronger negative evidence.
}

\paragraph{{Negative-to-positive transitions.}}
{
If a node previously held a negative q-value—indicating it only appeared in incorrect trajectories—but it now lies on a correct trajectory, we discard the accumulated negative value and overwrite it with the decayed positive reward. Intuitively, once a node is ever shown to lead to a valid solution, we reclassify it as potentially valuable rather than penalizing it for past failures.
}

\paragraph{{Positive-to-negative suppression.}}
{
If a node already has a positive q-value, meaning it has appeared at least once in a successful reasoning path, we do \emph{not} propagate new negative signals to it. A node proven capable of contributing to a correct solution should not be downgraded due to occasional failed attempts, which could arise from stochastic expansions or unrelated branches.
}

\paragraph{{Guaranteed invariants.}}
{
This asymmetric update mechanism enforces two key invariants:
\begin{enumerate}
\item Any node that has ever appeared on a correct trajectory retains a non-negative q-value.
\item Only nodes that have \emph{never} observed on a correct trajectory are allowed to accumulate stable negative values.
\end{enumerate}
Together, these invariants ensure that the search process robustly separates “potentially useful” nodes from those that consistently lead to dead ends, thereby stabilizing MCTS exploration and preventing destructive interference from noisy, incorrect rollouts.
}

\subsection{Evolution of Cached vs. Unsolved Problems During Adaptive Training}

{
To quantify how adaptive training reallocates computation over time, we track the ratio of cached problems (those with at least one verified correct trajectory discovered by MCTS) to unsolved problems (those with no verified correct trajectory).
Because Round 1 begins with an empty replay buffer, the cached proportion is initially $0{:}1$. As training progresses, MCTS increasingly cracks problems in the hard subset, causing cached solutions to accumulate round by round.
}

\begin{table}[h!]
\centering
\caption{Number of cached vs. unsolved problems during the 5-round adaptive training.}

\begin{tabular}{c|ccc} \toprule
\textbf{Round} & \textbf{\# Cached} & \textbf{\# Unsolved} & \textbf{Cached Ratio} \\ \midrule
1 & 0 & 13{,}658 & 0\% \\
2 & 765 & 12{,}893 & 5.6\% \\
3 & 1{,}452 & 9{,}423 & 13.4\%  \\
4 & 2{,}243 & 7{,}423 & 23.2\% \\
5 & 2{,}894 & 5{,}829 & 33.2\% \\ \bottomrule
\end{tabular}
\end{table}

{
The monotonic increase in cached solutions demonstrates that DeepSearch progressively transfers problems from the unsolved pool into the replay buffer as MCTS uncovers verifiable solution paths. By Round 5, one-third of the hard subset has at least one cached solution, while the unsolved set contracts to a concentrated collection of extremely difficult tail cases.
This behavior validates our description in Sec.~4.2: adaptive training naturally shifts computational effort toward the most challenging remaining problems, enabling more efficient use of MCTS and contributing to the strong sample efficiency shown in Table~2 and Figure~2.
}

\subsection{On the Choice of a Fixed Filtering Threshold $\delta$}

The filtering threshold $\delta = 25\%$ used in our adaptive training pipeline is a simple yet empirically effective heuristic. At each round, we compute each problem's success rate using \texttt{Pass1@K} with $K = 4$ direct rollouts and retain only those samples whose success rate falls below $\delta$ for inclusion in the next hard set. This mechanism filters out problems the model has effectively mastered while preserving sufficiently challenging cases that benefit most from training-time MCTS exploration.

Although one could design an adaptive $\delta$ schedule, such as one based on training progress or the evolving difficulty distribution, we intentionally avoid such complexity for two reasons. First, adaptive thresholding would introduce additional schedules and hyperparameters (e.g., functional forms across rounds and smoothing coefficients), complicating the design space and obscuring the specific contribution of MCTS-based exploration. Second, our primary aim is to demonstrate that \emph{even under a fixed, straightforward hard-sample selection rule}, integrating structured tree search into RLVR training already yields stable and significant gains.

Thus, while adaptive thresholding remains a promising direction aligned with curriculum-style learning, we prioritize methodological simplicity and clear attribution of improvements in this work.


\section{Computation profiling}

{To better understand the computational characteristics of DeepSearch, we profile the per-tree runtime of our final configuration ($\lambda_{1}=0.4$, $\lambda_{3}=0.01$), with components ranked from smallest to largest cost in Tab.~\ref{tab:mcts_breakdown}. The results show that all CPU-side operations—including prompt construction, terminal-state checking, rank synchronization, input preparation, node expansion, and output post-processing—collectively account for only a few milliseconds and are negligible relative to the total cost. In contrast, \emph{GPU-based policy inference} overwhelmingly dominates the runtime, with $t_{\text{generate}} = 1060.76$ ms comprising more than 99\% of the per-tree computation. This profiling confirms that the primary bottleneck lies in sequence generation rather than in MCTS logic, implying that further speedups will mainly come from optimizing model inference (e.g., batching, KV-cache reuse, speculative decoding) rather than from modifying the search algorithm's structure.}
\begin{table}[h]
\centering
\caption{\textbf{Computation time breakdown per MCTS tree} under the final search configuration ($\lambda_{1}=0.4$, $\lambda_{3}=0.01$). Components are ranked from smallest to largest cost. Times are reported in milliseconds as mean $\pm$ standard deviation.}
\label{tab:mcts_breakdown}
\resizebox{0.63\textwidth}{!}{
\begin{tabular}{l c} \toprule
\textbf{Component} & \textbf{Time} \\ \midrule
Output post-processing ($t_{\text{proc\_outputs}}$)    & $0.01 \pm 0.00$ \\
Rank synchronization ($t_{\text{sync\_ranks}}$)        & $0.07 \pm 0.01$ \\
Terminal-state checking ($t_{\text{check\_terminal}}$) & $0.09 \pm 0.02$ \\
Prompt construction ($t_{\text{get\_prompts}}$)       & $0.38 \pm 0.08$ \\
Node expansion ($t_{\text{expand\_node}}$)            & $0.96 \pm 0.18$ \\
Input preparation ($t_{\text{prep\_inputs}}$)         & $1.89 \pm 0.36$ \\ \midrule
\textbf{Policy inference / generation ($t_{\text{generate}}$)} 
& \textbf{$1060.76 \pm 106.59$} \\\bottomrule
\end{tabular}}
\end{table}


\section{Pseudocode of DeepSearch}

Algorithm~\ref{alg:DeepSearch} presents the complete DeepSearch framework, integrating MCTS-based exploration with adaptive training and replay buffer management. The algorithm operates via iterative refinement, progressively allocating computational resources to challenging problems while preserving previously solved solutions through intelligent caching. This integrated approach focuses on training-time exploration, enabling models to learn from both correct solutions and systematic exploration processes rather than relying solely on outcome-based supervision.

\begin{algorithm}[H]
\caption{DeepSearch with Global Frontier Selection and Iterative Filtering}
\label{alg:DeepSearch}
\begin{algorithmic}[1]
\Require Initial policy $\pi_{\theta^{(0)}}$, training set $\mathcal{D}_{\text{train}}$, verifier $\mathcal{V}$, filtering threshold $\delta$
\State Initialize $\mathcal{D}_{\text{hard}}^{(0)} \gets \{x \in \mathcal{D}_{\text{train}} \mid \texttt{Pass1@K}(x, \pi_{\theta^{(0)}}) < \delta^{(0)}\}$, $\mathcal{R}^{(0)} = \emptyset$
\For{training iteration $i = 0, 1, 2, \dots$}
    \State Initialize training trajectories $\mathcal{T}_{\text{train}}^{(i)} \gets \emptyset$
    \For{each batch $\mathcal{B}^{(i)} \in \mathcal{D}_{\text{hard}}^{(i)}$}
        \For{each problem $x \in \mathcal{B}^{(i)}$}
            \If{$(x, \mathbf{t}_{\text{cached}}) \in \mathcal{R}^{(i)}$} \Comment{Use cached solution}
                \State $\mathcal{T}_x \gets \{\mathbf{t}_{\text{cached}}\} \cup \text{DirectRollouts}(x, \beta)$
                \State $\mathcal{T}_{\text{train}}^{(i)} \gets \mathcal{T}_{\text{train}}^{(i)} \cup \mathcal{T}_x$
            \Else \Comment{Apply full MCTS search}
                \State \textcolor{red}{\textbf{MCTS Search:}}
                \State Initialize search tree $\mathcal{T}$ with root node $x$
                \For{rollout iteration $k = 1, 2, \dots$}
                    \If{$k = 1$} \Comment{Initial expansion from root}
                        \State Select root node $s^* = x$ for expansion
                    \Else
                        \State \textcolor{blue}{\textbf{Global Frontier Selection:}} 
                        \State Compute frontier set $\mathcal{F} = \{s \in \mathcal{T} \mid \xi(s) = 0, s \notin \mathcal{S}_{\text{end}}, d(s) < d_{\mathcal{T}}\}$
                        \State Compute frontier priority scores (Eq.~\ref{equ:frontier_score})
    
                        \State Select node $s^* = \arg\max_{s \in \mathcal{F}} F(s)$ for expansion
                    \EndIf
                    
                    \State \textcolor{blue}{\textbf{Local Expansion with UCT Selection:}}
                    \State Generate $n$ candidates $\{s_{j}\}_{j=1}^n \sim \pi_\theta(\cdot \mid o_{s^*})$ from $s^*$
                    \State Continue expansion until terminal nodes $\mathcal{S}_{\text{end}}^{(k)}$ are reached
                    
                    \State \textcolor{blue}{\textbf{Evaluation with Entropy-based Guidance}} 
                    \State Partition: $\mathcal{S}_{\text{correct}}^{(k)} = \{s \in \mathcal{S}_{\text{end}}^{(k)} \mid \mathcal{V}(s) = 1\}$, $\mathcal{S}_{\text{incorrect}}^{(k)} = \{s \in \mathcal{S}_{\text{end}}^{(k)} \mid \mathcal{V}(s) = 0\}$
                    \If{$|\mathcal{S}_{\text{correct}}^{(k)}| \geq 1$}
                        \State Extract trajectories $\mathbb{T}(x)$ from search tree $\mathcal{T}$
                        \State $\mathcal{T}_{\text{train}}^{(i)} \gets \mathcal{T}_{\text{train}}^{(i)} \cup \mathbb{T}(x)$
                    \Else
                        \State Select most confident negative: $s^*_{\text{neg}} = \arg\min_{s \in \mathcal{S}_{\text{incorrect}}^{(k)}} \bar{H}(\mathbf{t}(s))$
                    \EndIf
                    
                    \State \textcolor{blue}{\textbf{Heuristic Score Backup:}}
                    \State Select trajectory $\mathbf{t}^*$ (correct solution or $\mathbf{t}(s^*_{\text{neg}})$)
                    \State Assign terminal rewards (Eq.~\ref{equ:terminal_score})
                    \For{each node $s_j$ in $\mathbf{t}^*$}
                        \State Update Q-values using constrained backup rule (Eq.~\ref{equ:backup})
                    \EndFor
                \EndFor
            \EndIf
            
            \State \textcolor{blue}{\textbf{Replay Buffer Update:}}
            \If{MCTS found correct solutions but $\texttt{Pass1@K}(x, \pi_{\theta^{(i)}}) < \delta^{(i)}$}
                \State Add $(x, \mathbf{t}_{\text{correct}})$ to $\mathcal{R}^{(i+1)}$ for any correct $\mathbf{t}_{\text{correct}} \in \mathbb{T}(x)$
            \EndIf
        \EndFor
        \State \textcolor{blue}{\textbf{Policy Update:}}
        \State Update policy $\pi_{\theta^{(i+1)}}$ using Tree-GRPO objective on $\mathcal{T}_{\text{train}}^{(i)}$ (Eq.~\ref{equ:clip} and Eq.~\ref{equ:tree_loss})
    \EndFor
    
    \State Re-evaluate and filter: $\mathcal{D}_{\text{hard}}^{(i+1)} = \{x \in \mathcal{D}_{\text{hard}}^{(i)} \mid \texttt{Pass1@K}(x, \pi_{\theta^{(i+1)}}) < \delta^{(i+1)}\}$
\EndFor
\end{algorithmic}
\end{algorithm}

{\section{Qualitative Case Studies}}
\label{app:qual-case-study}

{To complement the benchmark results in the main text, we provide 6 representative cases between our \textbf{DeepSearch-1.5B} and \textbf{Nemotron-Research-Reasoning-Qwen-1.5B v2} on the standard math reasoning benchmarks.}
{Each case study is selected from a different dataset and illustrates typical differences in reasoning behavior, including hesitation patterns, error correction, and final answer reliability.}
{For each problem, we show the original question and the model-generated reasoning traces. For brevity, we present lightly abridged traces with minor formatting edits, while preserving the essential reasoning.}

{Across these 6 examples, DeepSearch-1.5B consistently reaches the correct solution with shorter, more stable reasoning chains, while the Nemotron-Research-Reasoning-Qwen-1.5B v2 often falls into long ``wait''-style hesitation loops, revisits the same partial calculations, and in several cases converges to an incorrect final answer despite extensive deliberation. For instance, in the symbolic-logic example from MATH, Nemotron-Research-Reasoning-Qwen-1.5B v2 emits over a thousand explicit hesitation markers, whereas DeepSearch-1.5B solves the problem with two orders of magnitude fewer such tokens while maintaining algebraic consistency and arriving at the correct value.}
{This pattern holds uniformly across different types of problems, highlighting that training-time exploration not only reduces unnecessary verbal dithering but also improves robustness of multi-step mathematical reasoning.}

\begin{casebox}[title=\textbf{MATH} (\emph{Symbol Logic / Constraint Satisfaction})]

\textbf{Question.}
The symbols $\triangle$, $\square$, $\diamond$, $\clubsuit$ represent four different integers from 1 to 9.
Using the equations below, what is the value of $\square$?
\begin{align*}
\triangle + \square &= \clubsuit \\
\triangle + \triangle &= \diamond + \diamond + \diamond + \diamond + \diamond \\
\triangle + \triangle &= \clubsuit + \diamond.
\end{align*}

\textbf{DeepSearch-1.5B.}
\begin{quote}
\small
Okay, so I have this problem where different symbols represent different integers from 1 to 9. The symbols are triangle ($\triangle$), square ($\square$), diamond ($\diamond$), and clubsuit ($\clubsuit$). The given equations are:

\begin{enumerate}
    \item \(\triangle + \square = \clubsuit\)
    \item \(\triangle + \triangle = \diamond + \diamond + \diamond + \diamond + \diamond\)
    \item \(\triangle + \triangle = \clubsuit + \diamond\)
\end{enumerate}

And I need to find the value of $\square$. Let's write these equations more clearly.

[...]

ng $a = 5$ and $c = 2$:
$10 = d + 2 \implies d = 8$

From equation 1: $a + b = d \implies 5 + b = 8 \implies b = 3$.

Thus, the values are:
- $a = 5$
- $b = 3$
- $c = 2$
- $d = 8$

All values are distinct integers from 1 to 9. Therefore, the value of $\square$ (which is $b$) is 3.

The value of \rlvrbold{$\square$ is $\rlvrbold{\boxed{3}}$}.
\end{quote}

\textbf{Nemotron-Research-Reasoning-Qwen-1.5B v2.}
\begin{quote}
\small
Okay, so I need to find the value of \(\square\) using the given equations. Let me read the problem again carefully.

We have four different integers: \(\triangle\), \(\square\), \(\diamond\), \(\clubsuit\), each representing distinct integers from 1 to 9. The equations given are:

\begin{enumerate}
    \item \(\triangle + \square = \clubsuit\)
    \item \(\triangle + \triangle = \diamond + \diamond + \diamond + \diamond + \diamond\)  (which is five diamonds)
    \item \(\triangle + \triangle = \clubsuit + \diamond\)
\end{enumerate}

[...]

\basebold{Wait} 5 +3 is 8? \basebold{Wait} 5 +3 is 8? \basebold{Wait} 5 +3 is 8? \basebold{Wait} 5 +3 is 8? \basebold{Wait} 5 +3 is 8? \basebold{Wait} 5 +3 is 8? \basebold{Wait} 5 +3 is 8? \basebold{Wait} 5 +3 is 8? \basebold{Wait} 5 +3 is 8? \basebold{Wait} 5 +3 is 8? \basebold{Wait} 5 +3 is 8? \basebold{Wait} 5 +3 is 8? \basebold{Wait} 5 +3 is 8? \basebold{Wait} 5 +3 is 8? \basebold{Wait} 5 +3 is 8? \basebold{Wait} 5 +3 is 8? \basebold{Wait} 5 +3 is 8? \basebold{Wait} 5 +3 is 8? \basebold{Wait} 5 +3 is 8? \basebold{Wait} 5 +3 is 8? \basebold{Wait} 5 +3 is 8? \basebold{Wait} 5 +3 is 8? \basebold{Wait} 5 +3 is 8? \basebold{Wait} 5 +3 is 8? \basebold{Wait} 5 +3 is 8? \basebold{Wait} 5 +3 is 8? \basebold{Wait} 5 +3 is 8? \basebold{Wait} 5 +3 is 8? \basebold{Wait} 5 +3 is 8? \basebold{Wait} 5 +3 is 8?

[...]

3. \(2a = 5\) where \(a = c\), solving gives \(a = 2.5\), but since \(a\) must be an integer, we find \(a = 5\).

Using substitutions and solving the system, we find the values consistent and integers. Ultimately, the value of \(\triangle\) is confirmed as 5.

Thus, the value of \(\basebold{\(\triangle\) is \boxed{5}}\).
\end{quote}

\end{casebox}

\begin{casebox}[title=\textbf{Olympiad} (\emph{Number Theory / Divisor Sum})]

\textbf{Question.}
Compute the integer $n$ such that $2009<n<3009$ and the sum of the odd positive divisors of $n$ is 1024.

\textbf{DeepSearch-1.5B.}
\begin{quote}
\small
Okay, so I need to find an integer \( n \) such that \( 2009 < n < 3009 \), and the sum of its odd positive divisors is 1024. Hmm. Let me break down the problem and think through it step by step.

First, let me recall that the sum of the divisors function is multiplicative, but since we're talking about odd divisors, maybe it's better to consider the prime factorization of \( n \).

[...]

$2009 < 2^k \times 651 < 3009$, we find \( k=2 \) gives \( n = 651 \times 4 = 2604 \), which lies in the required range.

Verifying the sum of divisors of the odd part (which are the divisors of 651) confirms it sums to 1024, and \( n = 2604 \) satisfies the conditions.

Thus, the integer \( n \) is \(\rlvrbold{\boxed{2604}}\).
\end{quote}

\textbf{Nemotron-Research-Reasoning-Qwen-1.5B v2.}
\begin{quote}
\small
Okay, so I need to find an integer \( n \) such that \( 2009 < n < 3009 \), and the sum of its odd positive divisors is 1024. Hmm, let's break this down.

First, the condition is about the sum of odd divisors being 1024. I know that the sum of divisors function is multiplicative and related to the prime factorization.

[...]

\basebold{Wait} Maybe I made an \basebold{error} earlier, and m = 1024 is allowed? \basebold{No,} m must be odd.

\basebold{Wait} \basebold{Wait} maybe k = 0: n = m = 889, between 2009? \basebold{No,} $889 < 2009$.

\basebold{Wait} problem might have no solution? But it says there exists n, so likely I made \basebold{mistake} somewhere.

\basebold{Wait} Re-examining $\sigma(m) = 1024$. Maybe $m = 511\times2 = 1022$ even, invalid.

\basebold{Wait} Maybe m = 512 is not allowed.

\basebold{Wait} \basebold{Wait} perhaps $m = 1024 - 1 = 1023$, $\sigma = 1536$.

\basebold{Wait} 1536/1.5 = 1024. No.

\basebold{Wait} \basebold{Wait} Alternatively, perhaps m = $2^{\text{something}}$? No.

\basebold{Wait} Maybe the answer is n = 1024? \basebold{No,} n must be between 2009 and 3009.

[...]

... but since problem states exists, perhaps $m = 889$ and $n = 1778$ invalid, but nearest is $n = 2047$, $\sum \sigma = 2160 \neq 1024$. Thus perhaps no solution.

But as per problem, let me take $n = 1024 \times something$, no. Wait final answer 2048? No, 2048 is even $\sigma = 1025 \neq 1024$.

Wait Let me accept $m = 889$, $k = 2$, $n = 3556$, so: \basebold{\boxed{3556}}

\end{quote}

\end{casebox}

\begin{casebox}[title=\textbf{AIME25} (\emph{Geometry/Polygon})]

\textbf{Question.}
Let $A_1 A_2 A_3 \ldots A_{11}$ be an $11$-sided non-convex simple polygon with the following properties:

\begin{itemize}
\item For every integer $2 \le i \le 10$,  the area of $\triangle A_i A_{1} A_{i+1}$  is equal to $1$.
\item For every integer $2 \le i \le 10$, $\cos(\angle A_i A_{1} A_{i+1}) = \frac{12}{13}$.
\item The perimeter of the $11$-gon $A_1 A_2 A_3 \dots A_{11}$  is equal to $20$.
\end{itemize}

Then $A_1 A_2 + A_1 A_{11} = \frac{m \sqrt{n} - p}{q}$ where $m, n, p$, and $q$ are positive integers, $n$ is not divisible by the square of any prime, and no prime divides all of $m, p$, and $q$. Find $m + n + p + q$.

\medskip

\textbf{DeepSearch-1.5B.}
\begin{quote}
\small
Okay, let's try to tackle this problem step by step. It seems a bit involved, but let's break it down.

First, let's summarize the given information:

We have an 11-sided non-convex simple polygon with specific properties:

1. For every integer \(2 \le i \le 10\), the area of triangle \(A_i A_1 A_{i+1}\) is equal to 1.
2. For each \(2 \le i \le 10\), the cosine of angle \(A_i A_1 A_{i+1}\) is 12/13.
3.

[...]

By introducing variables for the distances \(A_1A_i\) and exploiting the constant-area and constant-cosine constraints, we derive a system of equations tying together the edge lengths around the polygon. Matching these with the perimeter condition and carefully resolving the resulting quadratic relationships, we can express
\[
A_1 A_2 + A_1 A_{11} = d_2 + d_{11} = a + b = \frac{9\sqrt{5} - 1}{4}.
\]
Therefore, the values are \(m = 9\), \(n = 5\), \(p = 1\), \(q = 4\), leading to
\[
m + n + p + q = 19.
\]

\rlvrbold{\boxed{19}}
\end{quote}

\textbf{Nemotron-Research-Reasoning-Qwen-1.5B v2.}
\begin{quote}
\small
Okay, let's try to solve this problem. It's a bit complex, so I'll need to break it down into manageable steps.

First, let's list the given conditions:

We have an 11-sided non-convex simple polygon \( A_1 A_2 A_3 \ldots A_{11} \). The key properties:

1. For each integer \( 2 \le i \le 10 \), the area of triangle \( \triangle A_i A_1 A_{i+1} \) is equal to 1.
2.

[...]

Each pair \((x_i, x_{i+1})\) satisfies \(x_i x_{i+1} = \frac{26}{5}\). The indices range over the triangles with \(i\) from 2 to 10, giving
\[
x_2 x_3 = \frac{26}{5},\quad
x_3 x_4 = \frac{26}{5},\quad
\dots,\quad
x_{10} x_{11} = \frac{26}{5},
\]
where \(x_i = |A_i A_1|\).

[...]

\basebold{Wait} the problem states for each integer \(2 \le i \le 10\), triangle area and cosine condition. So \(i\) runs from 2 to 10.

Each triangle relates \(|A_i A_1| \cdot |A_{i+1} A_1| = \frac{26}{5}\).

Therefore:

For \(i = 2\): \(x_2 x_3 = \frac{26}{5}\),

\(i = 3\): \(x_3 x_4 = \frac{26}{5}\),

\dots

\(i = 10\): \(x_{10} x_{11} = \frac{26}{5}\).

[...]

\basebold{Wait} but the problem states ``\(A_1 A_2 + A_1 A_{11}\)'', so \(10/3\) is the answer? But the problem states it's expressed as \((m \sqrt{n} - p)/q\). Maybe I was \basebold{wrong} somewhere.

\basebold{Wait} but according to my deductions, it's \(10/3\), but let me verify.

\basebold{Wait} perhaps my area assumption was \basebold{wrong}? \basebold{Wait no,} the problem states \(\cos(\theta) = 12/13\), etc.

\basebold{Wait} let me double-check the perimeter condition.

\basebold{Wait} given \(x_2 + x_3 = 10/3\) and \(x_2 x_3 = 26/5\).

If we need \(A_1A_2 + A_1A_{11} = x_2 + x_3 = 10/3.\)

Expressed as \(10/3\), which is already in the form \((m \sqrt{n} - p)/q\)? \basebold{Wait} but it's rational. Maybe I made \basebold{mistake} in the earlier algebra.

[...]

Eventually, after struggling with the algebra and not reconciling the radical form requested by the problem, I \basebold{incorrectly} conclude
\[
\text{answer} = \frac{10}{3}.
\]

\basebold{The final answer is } \basebold{\(\boxed{\tfrac{10}{3}}\)}.
\end{quote}

\end{casebox}

\begin{casebox}[title=\textbf{AIME24} (\emph{Complex Number/Roots of Unity})]

\textbf{Question.}
Let $\omega \neq 1$ be a 13th root of unity. Find the remainder when
\[
\prod_{k=0}^{12}(2-2\omega^k+\omega^{2k})
\]
is divided by 1000.

\medskip

\textbf{DeepSearch-1.5B.}
\begin{quote}
\small
Okay, let me try to solve this problem step by step. The problem is: Let $\omega \neq 1$ be a 13th root of unity. Find the remainder when the product from $k=0$ to $12$ of $(2 - 2\omega^k + \omega^{2k})$ is divided by $1000$.

So, first, let me write down what is given. $\omega$ is a 13th root of unity, so $\omega^{13} = 1$ and $\omega^k \neq 1$ for $1 \le k \le 12$.

We need to compute the product
\[
P = \prod_{k=0}^{12} \bigl(2 - 2\omega^k + \omega^{2k}\bigr).
\]

[...]

Thus, evaluating a suitable polynomial at the roots, we use
\[
\prod_{k=0}^{12} (z_k - a) = -\bigl(a^{13} - 1\bigr),
\]
where $\{z_k\}$ are the 13th roots of unity. Setting up the expression carefully, we find
\[
A = (1 + i)^{13} = -64 - 64i,\qquad
B = (1 - i)^{13} = -64 + 64i,
\]
and the product evaluates to
\[
(A - 1)(B - 1) = 8321.
\]

Finally, the remainder modulo $1000$ is
\[
8321 \bmod 1000 = 321.
\]

Thus, the final answer is \rlvrbold{\boxed{321}}.
\end{quote}

\textbf{Nemotron-Research-Reasoning-Qwen-1.5B v2.}
\begin{quote}
\small
Okay, let's see. I need to find the remainder when the product
\[
\prod_{k=0}^{12}(2 - 2\omega^k + \omega^{2k})
\]
is divided by $1000$, where $\omega \neq 1$ is a 13th root of unity. Hmm.

First, let me note that $\omega$ is a 13th root of unity and satisfies $\omega^{13} = 1$.

[...]

Therefore, $\prod_{k=0}^{12} [(z -1) - i] = (z - 1 - i)^{13} - 1$? \basebold{Wait}, \basebold{wait} \basebold{wait}, \basebold{wait}: \basebold{Wait} the product $\prod_{k=0}^{12}[\cdots]$ is equal to $f(\text{some argument})$.

\basebold{Wait}, let me clarify:

Let me denote $f(w) = \prod_{k=0}^{12}(w - \omega^k) = w^{13} -1$. Then
\[
\prod_{k=0}^{12} \bigl( (z -1) - \omega^k \bigr) = (z -1)^{13} -1.
\]

[...]

Summarizing the (incorrect) reasoning:

2. Rewrite as $\omega^{2k} - 2\omega^k + 2 = (\omega^k - 1)^2 + 1$.

3. The product becomes $\prod_{k=0}^{12} \bigl[(\omega^k - 1)^2 + 1\bigr]$.

4. Recognize this (incorrectly) as evaluating a polynomial at roots of unity and conclude it equals $13$.

5. The resulting product modulo $1000$ is then taken to be $13$.

Thus, the answer is \basebold{\boxed{13}}.
\end{quote}

\end{casebox}

\begin{casebox}[title=\textbf{AMC23} (\emph{Complex Number Optimization})]

\textbf{Question.}
Consider the set of complex numbers $z$ satisfying $|1+z+z^{2}|=4$. The maximum value of the imaginary part of $z$ can be written in the form $\tfrac{\sqrt{m}}{n}$, where $m$ and $n$ are relatively prime positive integers. What is $m+n$?

\medskip

\textbf{DeepSearch-1.5B.}
\begin{quote}
\small
Okay, let's try to solve this problem. The problem states: Consider the set of complex numbers \( z \) satisfying \( |1 + z + z^2| = 4 \). We need the maximum value of the imaginary part of \( z \), express it as \( \frac{\sqrt{m}}{n} \) with \( m, n \) coprime positive integers, and find \( m + n \).

Alright, let's denote \( z \) as a complex number. Let me set \( z = x + yi \), where \( x \) and \( y \) are real numbers.

[...]

By setting \( k = 0 \) and solving, we find the maximum imaginary part
\[
y = \frac{\sqrt{19}}{2}.
\]
Thus, the maximum imaginary part is \( \frac{\sqrt{19}}{2} \), leading to \( m = 19 \) and \( n = 2 \). Therefore,
\[
m + n = 21.
\]

\rlvrbold{\boxed{21}}
\end{quote}

\textbf{Nemotron-Research-Reasoning-Qwen-1.5B v2.}
\begin{quote}
\small
Okay, so I need to find the maximum value of the imaginary part of a complex number $z$ such that $|1 + z + z^{2}| = 4$. Then express this maximum value as $\sqrt{m}/n$ where $m$ and $n$ are coprime positive integers and find $m + n$.

First, let me consider $z$ as a complex number. Let me denote $z = x + yi$ where $x$ and $y$ are real numbers. Then I can express $1 + z + z^{2}$ in terms of $x$ and $y$ and then use the modulus condition.

[...]

Thus maximum imaginary part is $\sqrt{39}/6$, so $m=39$, $n=6$? Check $\gcd(39,6)=3$, so reduce: $39/3=13$, $6/3=2$, so $13/2$ under the square root? \basebold{Wait} no:

\basebold{Wait} $\sqrt{39}/6$ is already simplest. \basebold{Wait} \basebold{wait}:

\basebold{Wait} $y^{2} = 13/12$, so $y = \sqrt{13/12} = \sqrt{13}/(2\sqrt{3}) = \sqrt{39}/6$. Yes, rationalize:
\[
\frac{\sqrt{13}}{2\sqrt{3}} = \frac{\sqrt{13}\sqrt{3}}{2\cdot 3} = \frac{\sqrt{39}}{6}.
\]

Thus maximum value is $\sqrt{39}/6$. So $m=39$, $n=6$. But check if coprime: $39$ and $6$ have gcd $3$, reduce:

$39/3=13$, $6/3=2$, so $\sqrt{13/2}\cdot \text{something?}$ \basebold{Wait no,} $\sqrt{39}/6$ is reduced.

\basebold{Wait} $\sqrt{39}$

[...]

The maximum value of the imaginary part of \( z \) given the condition \( |1 + z + z^2| = 4 \) is (incorrectly) found to be simplified to
\[
\frac{\sqrt{13}}{2},
\]
so \( m = 13 \) and \( n = 2 \), leading to \( m + n = 15 \).

Thus, the final answer is \basebold{\boxed{15}}.

\end{quote}

\end{casebox}

\begin{casebox}[title=\textbf{Minerva} (\emph{Astrophysics/Spectroscopy})]

\textbf{Question.}
Preamble: A spectrum is taken of a single star (i.e., one not in a binary). Among the observed spectral lines is one from oxygen whose rest wavelength is $5007 \,\AA$. The Doppler shifted oxygen line from this star is observed to be at a wavelength of $5012 \,\AA$. The star is also observed to have a proper motion, $\mu$, of 1 arc second per year (which corresponds to $\sim 1.5 \times 10^{-13}$ radians per second of time). It is located at a distance of $60 \,\mathrm{pc}$ from the Earth. Take the speed of light to be $3 \times 10^8$ meters per second.

What is the component of the star's velocity parallel to its vector to the Earth (in kilometers per second)?

\medskip

\textbf{DeepSearch-1.5B.}
\begin{quote}
\small
Okay, let me try to figure this out step by step.

First, let's parse the problem.

We have a star with a single spectrum observed. One of its spectral lines (oxygen) has its rest wavelength at $5007$ Angstroms (Å). When observed, this line is shifted to $5012$ Å. We also know the proper motion $\mu$ is 1 arcsecond per year, distance is 60 parsecs, speed of light is given as $3\times 10^8$ m/s.

[...]

Using the Doppler formula for non-relativistic velocities,
\[
v_r \approx \frac{\Delta\lambda}{\lambda} c
= \frac{5012 - 5007}{5007} \cdot 3\times 10^8 \,\text{m/s}
\approx 2.9958\times 10^5 \,\text{m/s}
= 299.58 \,\text{km/s}
\approx 300 \,\text{km/s}.
\]

Thus, the component of the star's velocity parallel to its line of sight is approximately
\[
\rlvrbold{\boxed{300 \,\text{km/s}}}.
\]

\rlvrbold{Final Answer:}
The radial velocity component is approximately \rlvrbold{\boxed{300}} km/s.
\end{quote}

\textbf{Nemotron-Research-Reasoning-Qwen-1.5B v2.}
\begin{quote}
\small
Okay, let's see, the problem is about finding the component of the star's velocity parallel to the vector towards Earth. Hmm.

First, let me read the problem again:

We have a star with a rest wavelength of an oxygen line at $5007$ Angstroms. A Doppler shifted line is observed at $5012$ Angstroms. The star has a proper motion $\mu$ of 1 arcsecond per year, and distance is 60 pc. Speed of light is given as $3\times 10^8$ m/s.

[...]

So $2.844\times 10^{-4}$ m/s $\times 10^{-3}$ km/(m/s) \basebold{wait} no:

\basebold{Wait} $2.844\times 10^{-4}$ m/s $= 2.844\times 10^{-4} \cdot (1 \text{ km}/10^3 \text{ m})$ * (1e3 s/km? \basebold{Wait} let me think:

\basebold{Wait} $1$ m/s $= 0.001$ km/s? \basebold{no,} \basebold{wait} $1$ m/s $= (1/1000)\,\text{km}/\text{s} = 0.001$ km/s.

\basebold{Wait} so $2.844\times 10^{-4}$ m/s $= 2.844\times 10^{-4} \cdot 0.001$ km/s? \basebold{Wait} \basebold{wait}:

\basebold{Wait} $1$ m/s $= 1$ m/s $= (1/1000\,\text{km})/\text{s} = 0.001$ km/s.

Thus, $2.844\times 10^{-4}$ m/s is mishandled, and the final radial velocity estimate drifts.

[...]

After approximate reasoning, I end up with a value around $284$ km/s.

Thus the answer is approximately $284$ km/s.

The component of the star's velocity parallel to its direction is approximately $284$ km/s.

\basebold{Final Answer}

The component of the star's velocity is \basebold{\boxed{284 \text{ km/s}}}.
\end{quote}

\end{casebox}

\section*{Limitations and Future Work}

A critical next step is to extend DeepSearch beyond mathematical reasoning to domains with distinct verification mechanisms. This includes developing approximate verifiers for subjective tasks, exploring human-in-the-loop validation for complex reasoning chains, and investigating transfer-learning approaches that leverage mathematical-reasoning capabilities for broader problem-solving tasks. Research into domain-agnostic reward functions and verification strategies could significantly expand the framework's applicability.

Another promising direction is learning the MCTS components themselves. In this work, the frontier priority function and expansion width are fixed heuristics selected through systematic offline experiments, as they offer stable performance without introducing additional optimization complexity. Making these components learnable would require a fundamentally different training paradigm, such as an AlphaZero-style controller that jointly optimizes search and policy. Exploring end-to-end trainable search guidance remains an exciting opportunity for further enhancing training-time exploration.

\end{document}

%% file: math_commands.tex
\usepackage{amsmath,amsfonts,bm}

\def\eqref#1{equation~\ref{#1}}

\def\1{\bm{1}}

\DeclareMathAlphabet{\mathsfit}{\encodingdefault}{\sfdefault}{m}{sl}
\SetMathAlphabet{\mathsfit}{bold}{\encodingdefault}{\sfdefault}{bx}{n}



%% file: cite.bib
@article{wang2023math,
  title={Math-shepherd: Verify and reinforce llms step-by-step without human annotations},
  author={Wang, Peiyi and Li, Lei and Shao, Zhihong and Xu, RX and Dai, Damai and Li, Yifei and Chen, Deli and Wu, Yu and Sui, Zhifang},
  journal={arXiv preprint arXiv:2312.08935},
  year={2023}
}

@article{chen2024alphamath,
  title={Alphamath almost zero: process supervision without process},
  author={Chen, Guoxin and Liao, Minpeng and Li, Chengxi and Fan, Kai},
  journal={Advances in Neural Information Processing Systems},
  volume={37},
  pages={27689--27724},
  year={2024}
}

@inproceedings{lightman2023let,
  title={Let's verify step by step},
  author={Lightman, Hunter and Kosaraju, Vineet and Burda, Yuri and Edwards, Harrison and Baker, Bowen and Lee, Teddy and Leike, Jan and Schulman, John and Sutskever, Ilya and Cobbe, Karl},
  booktitle={The Twelfth International Conference on Learning Representations},
  year={2023}
}

@article{zhang2024llama,
  title={Llama-berry: Pairwise optimization for o1-like olympiad-level mathematical reasoning},
  author={Zhang, Di and Wu, Jianbo and Lei, Jingdi and Che, Tong and Li, Jiatong and Xie, Tong and Huang, Xiaoshui and Zhang, Shufei and Pavone, Marco and Li, Yuqiang and others},
  journal={arXiv preprint arXiv:2410.02884},
  year={2024}
}

@article{jaech2024openai,
  title={Openai o1 system card},
  author={Jaech, Aaron and Kalai, Adam and Lerer, Adam and Richardson, Adam and El-Kishky, Ahmed and Low, Aiden and Helyar, Alec and Madry, Aleksander and Beutel, Alex and Carney, Alex and others},
  journal={arXiv preprint arXiv:2412.16720},
  year={2024}
}

@article{wan2025srpo,
  title={Srpo: Enhancing multimodal llm reasoning via reflection-aware reinforcement learning},
  author={Wan, Zhongwei and Dou, Zhihao and Liu, Che and Zhang, Yu and Cui, Dongfei and Zhao, Qinjian and Shen, Hui and Xiong, Jing and Xin, Yi and Jiang, Yifan and others},
  journal={arXiv preprint arXiv:2506.01713},
  year={2025}
}

@article{wang2025rlver,
  title={Rlver: Reinforcement learning with verifiable emotion rewards for empathetic agents},
  author={Wang, Peisong and Ma, Ruotian and Zhang, Bang and Chen, Xingyu and He, Zhiwei and Luo, Kang and Lv, Qingsong and Jiang, Qingxuan and Xie, Zheng and Wang, Shanyi and others},
  journal={arXiv preprint arXiv:2507.03112},
  year={2025}
}

@article{gunjal2025rubrics,
  title={Rubrics as rewards: Reinforcement learning beyond verifiable domains},
  author={Gunjal, Anisha and Wang, Anthony and Lau, Elaine and Nath, Vaskar and Liu, Bing and Hendryx, Sean},
  journal={arXiv preprint arXiv:2507.17746},
  year={2025}
}

@article{su2025crossing,
  title={Crossing the Reward Bridge: Expanding RL with Verifiable Rewards Across Diverse Domains},
  author={Su, Yi and Yu, Dian and Song, Linfeng and Li, Juntao and Mi, Haitao and Tu, Zhaopeng and Zhang, Min and Yu, Dong},
  journal={arXiv preprint arXiv:2503.23829},
  year={2025}
}

@article{yue2025does,
  title={Does reinforcement learning really incentivize reasoning capacity in llms beyond the base model?},
  author={Yue, Yang and Chen, Zhiqi and Lu, Rui and Zhao, Andrew and Wang, Zhaokai and Song, Shiji and Huang, Gao},
  journal={arXiv preprint arXiv:2504.13837},
  year={2025}
}

@article{wang2025reinforcement,
  title={Reinforcement learning for reasoning in large language models with one training example},
  author={Wang, Yiping and Yang, Qing and Zeng, Zhiyuan and Ren, Liliang and Liu, Liyuan and Peng, Baolin and Cheng, Hao and He, Xuehai and Wang, Kuan and Gao, Jianfeng and others},
  journal={arXiv preprint arXiv:2504.20571},
  year={2025}
}

@article{vagadia2024phyplan,
  title={Phyplan: Compositional and adaptive physical task reasoning with physics-informed skill networks for robot manipulators},
  author={Vagadia, Harshil and Chopra, Mudit and Barnawal, Abhinav and Banerjee, Tamajit and Tuli, Shreshth and Chakraborty, Souvik and Paul, Rohan},
  journal={arXiv preprint arXiv:2402.15767},
  year={2024}
}

@article{yang2023integrated,
  title={An integrated framework integrating monte carlo tree search and supervised learning for train timetabling problem},
  author={Yang, Feiyu},
  journal={arXiv preprint arXiv:2311.00971},
  year={2023}
}

@inproceedings{pitanov2023monte,
  title={Monte-carlo tree search for multi-agent pathfinding: Preliminary results},
  author={Pitanov, Yelisey and Skrynnik, Alexey and Andreychuk, Anton and Yakovlev, Konstantin and Panov, Aleksandr},
  booktitle={International Conference on Hybrid Artificial Intelligence Systems},
  pages={649--660},
  year={2023},
  organization={Springer}
}

@article{wang2022self,
  title={Self-consistency improves chain of thought reasoning in language models},
  author={Wang, Xuezhi and Wei, Jason and Schuurmans, Dale and Le, Quoc and Chi, Ed and Narang, Sharan and Chowdhery, Aakanksha and Zhou, Denny},
  journal={arXiv preprint arXiv:2203.11171},
  year={2022}
}

@article{qi2024mutual,
  title={Mutual reasoning makes smaller llms stronger problem-solvers},
  author={Qi, Zhenting and Ma, Mingyuan and Xu, Jiahang and Zhang, Li Lyna and Yang, Fan and Yang, Mao},
  journal={arXiv preprint arXiv:2408.06195},
  year={2024}
}

@article{zhang2024accessing,
  title={Accessing gpt-4 level mathematical olympiad solutions via monte carlo tree self-refine with llama-3 8b},
  author={Zhang, Di and Huang, Xiaoshui and Zhou, Dongzhan and Li, Yuqiang and Ouyang, Wanli},
  journal={arXiv preprint arXiv:2406.07394},
  year={2024}
}

@article{wu2024inference,
  title={Inference scaling laws: An empirical analysis of compute-optimal inference for problem-solving with language models},
  author={Wu, Yangzhen and Sun, Zhiqing and Li, Shanda and Welleck, Sean and Yang, Yiming},
  journal={arXiv preprint arXiv:2408.00724},
  year={2024}
}

@article{snell2024scaling,
  title={Scaling llm test-time compute optimally can be more effective than scaling model parameters},
  author={Snell, Charlie and Lee, Jaehoon and Xu, Kelvin and Kumar, Aviral},
  journal={arXiv preprint arXiv:2408.03314},
  year={2024}
}

@article{hochlehnert2025sober,
  title={A sober look at progress in language model reasoning: Pitfalls and paths to reproducibility},
  author={Hochlehnert, Andreas and Bhatnagar, Hardik and Udandarao, Vishaal and Albanie, Samuel and Prabhu, Ameya and Bethge, Matthias},
  journal={arXiv preprint arXiv:2504.07086},
  year={2025}
}

@article{liu2025prorl,
  title={Prorl: Prolonged reinforcement learning expands reasoning boundaries in large language models},
  author={Liu, Mingjie and Diao, Shizhe and Lu, Ximing and Hu, Jian and Dong, Xin and Choi, Yejin and Kautz, Jan and Dong, Yi},
  journal={arXiv preprint arXiv:2505.24864},
  year={2025}
}

@article{sheng2024hybridflow,
  title   = {HybridFlow: A Flexible and Efficient RLHF Framework},
  author  = {Guangming Sheng and Chi Zhang and Zilingfeng Ye and Xibin Wu and Wang Zhang and Ru Zhang and Yanghua Peng and Haibin Lin and Chuan Wu},
  year    = {2024},
  journal = {arXiv preprint arXiv: 2409.19256}
}

@article{metropolis1949monte,
  title={The monte carlo method},
  author={Metropolis, Nicholas and Ulam, Stanislaw},
  journal={Journal of the American statistical association},
  volume={44},
  number={247},
  pages={335--341},
  year={1949},
  publisher={Taylor \& Francis}
}

@article{lyu2025exploring,
  title={Exploring the limit of outcome reward for learning mathematical reasoning},
  author={Lyu, Chengqi and Gao, Songyang and Gu, Yuzhe and Zhang, Wenwei and Gao, Jianfei and Liu, Kuikun and Wang, Ziyi and Li, Shuaibin and Zhao, Qian and Huang, Haian and others},
  journal={arXiv preprint arXiv:2502.06781},
  year={2025}
}

@article{he2025response,
  title={Response-Level Rewards Are All You Need for Online Reinforcement Learning in LLMs: A Mathematical Perspective},
  author={He, Shenghua and Xia, Tian and Zhou, Xuan and Wei, Hui},
  journal={arXiv preprint arXiv:2506.02553},
  year={2025}
}

@article{liu2025part,
  title={Part i: Tricks or traps? a deep dive into rl for llm reasoning},
  author={Liu, Zihe and Liu, Jiashun and He, Yancheng and Wang, Weixun and Liu, Jiaheng and Pan, Ling and Hu, Xinyu and Xiong, Shaopan and Huang, Ju and Hu, Jian and others},
  journal={arXiv preprint arXiv:2508.08221},
  year={2025}
}

@article{bi2024forest,
  title={Forest-of-thought: Scaling test-time compute for enhancing llm reasoning},
  author={Bi, Zhenni and Han, Kai and Liu, Chuanjian and Tang, Yehui and Wang, Yunhe},
  journal={arXiv preprint arXiv:2412.09078},
  year={2024}
}

@article{liu2025understanding,
  title={Understanding r1-zero-like training: A critical perspective},
  author={Liu, Zichen and Chen, Changyu and Li, Wenjun and Qi, Penghui and Pang, Tianyu and Du, Chao and Lee, Wee Sun and Lin, Min},
  journal={arXiv preprint arXiv:2503.20783},
  year={2025}
}

@article{he2024olympiadbench,
  title={Olympiadbench: A challenging benchmark for promoting agi with olympiad-level bilingual multimodal scientific problems},
  author={He, Chaoqun and Luo, Renjie and Bai, Yuzhuo and Hu, Shengding and Thai, Zhen Leng and Shen, Junhao and Hu, Jinyi and Han, Xu and Huang, Yujie and Zhang, Yuxiang and others},
  journal={arXiv preprint arXiv:2402.14008},
  year={2024}
}

@article{lewkowycz2022solving,
  title={Solving quantitative reasoning problems with language models},
  author={Lewkowycz, Aitor and Andreassen, Anders and Dohan, David and Dyer, Ethan and Michalewski, Henryk and Ramasesh, Vinay and Slone, Ambrose and Anil, Cem and Schlag, Imanol and Gutman-Solo, Theo and others},
  journal={Advances in neural information processing systems},
  volume={35},
  pages={3843--3857},
  year={2022}
}

@article{hendrycks2021measuring,
  title={Measuring mathematical problem solving with the math dataset},
  author={Hendrycks, Dan and Burns, Collin and Kadavath, Saurav and Arora, Akul and Basart, Steven and Tang, Eric and Song, Dawn and Steinhardt, Jacob},
  journal={arXiv preprint arXiv:2103.03874},
  year={2021}
}

@article{bereket2025uncalibrated,
  title={Uncalibrated Reasoning: GRPO Induces Overconfidence for Stochastic Outcomes},
  author={Bereket, Michael and Leskovec, Jure},
  journal={arXiv preprint arXiv:2508.11800},
  year={2025}
}

@article{fawzi2022discovering,
  title={Discovering faster matrix multiplication algorithms with reinforcement learning},
  author={Fawzi, Alhussein and Balog, Matej and Huang, Aja and Hubert, Thomas and Romera-Paredes, Bernardino and Barekatain, Mohammadamin and Novikov, Alexander and R. Ruiz, Francisco J and Schrittwieser, Julian and Swirszcz, Grzegorz and others},
  journal={Nature},
  volume={610},
  number={7930},
  pages={47--53},
  year={2022},
  publisher={Nature Publishing Group UK London}
}

@article{best2019dec,
  title={Dec-MCTS: Decentralized planning for multi-robot active perception},
  author={Best, Graeme and Cliff, Oliver M and Patten, Timothy and Mettu, Ramgopal R and Fitch, Robert},
  journal={The International Journal of Robotics Research},
  volume={38},
  number={2-3},
  pages={316--337},
  year={2019},
  publisher={SAGE Publications Sage UK: London, England}
}

@article{lample2022hypertree,
  title={Hypertree proof search for neural theorem proving},
  author={Lample, Guillaume and Lacroix, Timothee and Lachaux, Marie-Anne and Rodriguez, Aurelien and Hayat, Amaury and Lavril, Thibaut and Ebner, Gabriel and Martinet, Xavier},
  journal={Advances in neural information processing systems},
  volume={35},
  pages={26337--26349},
  year={2022}
}

@article{ye2021mastering,
  title={Mastering atari games with limited data},
  author={Ye, Weirui and Liu, Shaohuai and Kurutach, Thanard and Abbeel, Pieter and Gao, Yang},
  journal={Advances in neural information processing systems},
  volume={34},
  pages={25476--25488},
  year={2021}
}

@article{dam2022monte,
  title={Monte-carlo robot path planning},
  author={Dam, Tuan and Chalvatzaki, Georgia and Peters, Jan and Pajarinen, Joni},
  journal={IEEE Robotics and Automation Letters},
  volume={7},
  number={4},
  pages={11213--11220},
  year={2022},
  publisher={IEEE}
}

@article{guo2025deepseek,
  title={DeepSeek-R1 incentivizes reasoning in LLMs through reinforcement learning},
  author={Guo, Daya and Yang, Dejian and Zhang, Haowei and Song, Junxiao and Wang, Peiyi and Zhu, Qihao and Xu, Runxin and Zhang, Ruoyu and Ma, Shirong and Bi, Xiao and others},
  journal={Nature},
  volume={645},
  number={8081},
  pages={633--638},
  year={2025},
  publisher={Nature Publishing Group}
}

@article{he2025skywork,
  title={Skywork open reasoner 1 technical report},
  author={He, Jujie and Liu, Jiacai and Liu, Chris Yuhao and Yan, Rui and Wang, Chaojie and Cheng, Peng and Zhang, Xiaoyu and Zhang, Fuxiang and Xu, Jiacheng and Shen, Wei and others},
  journal={arXiv preprint arXiv:2505.22312},
  year={2025}
}

@article{kemmerling2023beyond,
  title={Beyond games: a systematic review of neural Monte Carlo tree search applications},
  author={Kemmerling, Marco and L{\"u}tticke, Daniel and Schmitt, Robert H},
  journal={arXiv preprint arXiv:2303.08060},
  year={2023}
}

@article{luo2025deepcoder,
  title={Deepcoder: A fully open-source 14b coder at o3-mini level},
  author={Luo, Michael and Tan, Sijun and Huang, Roy and Patel, Ameen and Ariyak, Alpay and Wu, Qingyang and Shi, Xiaoxiang and Xin, Rachel and Cai, Colin and Weber, Maurice and others},
  journal={Notion Blog},
  year={2025}
}

@inproceedings{kocsis2006bandit,
  title={Bandit based monte-carlo planning},
  author={Kocsis, Levente and Szepesv{\'a}ri, Csaba},
  booktitle={European conference on machine learning},
  pages={282--293},
  year={2006},
  organization={Springer}
}

@article{silver2016mastering,
  title={Mastering the game of Go with deep neural networks and tree search},
  author={Silver, David and Huang, Aja and Maddison, Chris J and Guez, Arthur and Sifre, Laurent and Van Den Driessche, George and Schrittwieser, Julian and Antonoglou, Ioannis and Panneershelvam, Veda and Lanctot, Marc and others},
  journal={nature},
  volume={529},
  number={7587},
  pages={484--489},
  year={2016},
  publisher={Nature Publishing Group}
}

@article{yao2023tree,
  title={Tree of thoughts: Deliberate problem solving with large language models},
  author={Yao, Shunyu and Yu, Dian and Zhao, Jeffrey and Shafran, Izhak and Griffiths, Tom and Cao, Yuan and Narasimhan, Karthik},
  journal={Advances in neural information processing systems},
  volume={36},
  pages={11809--11822},
  year={2023}
}

@inproceedings{li2023making,
  title={Making language models better reasoners with step-aware verifier},
  author={Li, Yifei and Lin, Zeqi and Zhang, Shizhuo and Fu, Qiang and Chen, Bei and Lou, Jian-Guang and Chen, Weizhu},
  booktitle={Proceedings of the 61st Annual Meeting of the Association for Computational Linguistics (Volume 1: Long Papers)},
  pages={5315--5333},
  year={2023}
}

@article{wu2025invisible,
  title={The Invisible Leash: Why RLVR May Not Escape Its Origin},
  author={Wu, Fang and Xuan, Weihao and Lu, Ximing and Harchaoui, Zaid and Choi, Yejin},
  journal={arXiv preprint arXiv:2507.14843},
  year={2025}
}

@article{zhang2024rest,
  title={Rest-mcts*: Llm self-training via process reward guided tree search},
  author={Zhang, Dan and Zhoubian, Sining and Hu, Ziniu and Yue, Yisong and Dong, Yuxiao and Tang, Jie},
  journal={Advances in Neural Information Processing Systems},
  volume={37},
  pages={64735--64772},
  year={2024}
}

@article{he2025deepmath,
  title={Deepmath-103k: A large-scale, challenging, decontaminated, and verifiable mathematical dataset for advancing reasoning},
  author={He, Zhiwei and Liang, Tian and Xu, Jiahao and Liu, Qiuzhi and Chen, Xingyu and Wang, Yue and Song, Linfeng and Yu, Dian and Liang, Zhenwen and Wang, Wenxuan and others},
  journal={arXiv preprint arXiv:2504.11456},
  year={2025}
}

@article{bai2025intern,
  title={Intern-S1: A Scientific Multimodal Foundation Model},
  author={Bai, Lei and Cai, Zhongrui and Cao, Maosong and Cao, Weihan and Chen, Chiyu and Chen, Haojiong and Chen, Kai and Chen, Pengcheng and Chen, Ying and Chen, Yongkang and others},
  journal={arXiv preprint arXiv:2508.15763},
  year={2025}
}

@article{yu2025dapo,
  title={Dapo: An open-source llm reinforcement learning system at scale},
  author={Yu, Qiying and Zhang, Zheng and Zhu, Ruofei and Yuan, Yufeng and Zuo, Xiaochen and Yue, Yu and Dai, Weinan and Fan, Tiantian and Liu, Gaohong and Liu, Lingjun and others},
  journal={arXiv preprint arXiv:2503.14476},
  year={2025}
}

@article{Slow_Thinking_with_LLMs_3_Preview,
  title={STILL-3-1.5B-preview: Enhancing Slow Thinking Abilities of Small Models through Reinforcement Learning
},
  author={RUCAIBox STILL Team},
  url={https://github.com/RUCAIBox/Slow_Thinking_with_LLMs},
  year={2025}
}

@article{guan2025rstar,
  title={rStar-Math: Small LLMs Can Master Math Reasoning with Self-Evolved Deep Thinking},
  author={Guan, Xinyu and Zhang, Li Lyna and Liu, Yifei and Shang, Ning and Sun, Youran and Zhu, Yi and Yang, Fan and Yang, Mao},
  journal={arXiv preprint arXiv:2501.04519},
  year={2025}
}

@article{chu2025gpg,
  title={GPG: A Simple and Strong Reinforcement Learning Baseline for Model Reasoning},
  author={Chu, Xiangxiang and Huang, Hailang and Zhang, Xiao and Wei, Fei and Wang, Yong},
  journal={arXiv preprint arXiv:2504.02546},
  year={2025}
}

@article{dang2025reinforcement,
  title={Reinforcement Learning for Reasoning in Small LLMs: What Works and What Doesn't},
  author={Dang, Quy-Anh and Ngo, Chris},
  journal={arXiv preprint arXiv:2503.16219},
  year={2025}
}

@article{xia2025agent0,
  title={Agent0: Unleashing self-evolving agents from zero data via tool-integrated reasoning},
  author={Xia, Peng and Zeng, Kaide and Liu, Jiaqi and Qin, Can and Wu, Fang and Zhou, Yiyang and Xiong, Caiming and Yao, Huaxiu},
  journal={arXiv preprint arXiv:2511.16043},
  year={2025}
}

@article{wang2026reasoning,
  title={Why Reasoning Fails to Plan: A Planning-Centric Analysis of Long-Horizon Decision Making in LLM Agents},
  author={Wang, Zehong and Wu, Fang and Wang, Hongru and Tang, Xiangru and Li, Bolian and Yin, Zhenfei and Ma, Yijun and Li, Yiyang and Sun, Weixiang and Chen, Xiusi and others},
  journal={arXiv preprint arXiv:2601.22311},
  year={2026}
}

@article{tu2025position,
  title={Position: The Hidden Costs and Measurement Gaps of Reinforcement Learning with Verifiable Rewards},
  author={Tu, Aaron and Xuan, Weihao and Qi, Heli and Huang, Xu and Zeng, Qingcheng and Talaei, Shayan and Xiao, Yijia and Xia, Peng and Tang, Xiangru and Zhuang, Yuchen and others},
  journal={arXiv preprint arXiv:2509.21882},
  year={2025}
}

@inproceedings{wu2025large,
  title={Large language models are good relational learners},
  author={Wu, Fang and Dwivedi, Vijay Prakash and Leskovec, Jure},
  booktitle={Proceedings of the 63rd Annual Meeting of the Association for Computational Linguistics (Volume 1: Long Papers)},
  pages={7835--7854},
  year={2025}
}

@misc{deepseekai2025deepseekr1incentivizingreasoningcapability,
      title={DeepSeek-R1: Incentivizing Reasoning Capability in LLMs via Reinforcement Learning}, 
      author={DeepSeek-AI},
      year={2025},
      eprint={2501.12948},
      archivePrefix={arXiv},
      primaryClass={cs.CL},
      url={https://arxiv.org/abs/2501.12948}, 
}

@article{zhihong2024deepseekmath,
    title        = {{DeepSeekMath: Pushing the Limits of Mathematical Reasoning in Open Language Models}},
    author       = {Zhihong Shao and Peiyi Wang and Qihao Zhu and Runxin Xu and Junxiao Song and Mingchuan Zhang and Y. K. Li and Y. Wu and Daya Guo},
    year         = 2024,
    eprint       = {arXiv:2402.03300},
}

@article{yang2024qwen25mathtechnicalreportmathematical,
  title={Qwen2.5-Math Technical Report: Toward Mathematical Expert Model via Self-Improvement}, 
  author={An Yang and Beichen Zhang and Binyuan Hui and Bofei Gao and Bowen Yu and Chengpeng Li and Dayiheng Liu and Jianhong Tu and Jingren Zhou and Junyang Lin and Keming Lu and Mingfeng Xue and Runji Lin and Tianyu Liu and Xingzhang Ren and Zhenru Zhang},
  journal={arXiv preprint arXiv:2409.12122},
  year={2024}
}

@misc{deepscaler2025,
  title={DeepScaleR: Surpassing O1-Preview with a 1.5B Model by Scaling RL},
  author={Michael Luo and Sijun Tan and Justin Wong and Xiaoxiang Shi and William Y. Tang and Manan Roongta and Colin Cai and Jeffrey Luo and Li Erran Li and Raluca Ada Popa and Ion Stoica},
  year={2025},
  note={Notion Blog},
  year={2025}
}
